\documentclass{article}
\usepackage{graphicx}

 \usepackage[dblblindworkshop, final]{neurips_2025}

\usepackage[utf8]{inputenc} 
\usepackage[T1]{fontenc}    
\usepackage{hyperref}       
\usepackage{url}            
\usepackage{booktabs}       
\usepackage{amsfonts}       
\usepackage{nicefrac}       
\usepackage{microtype}      
\usepackage{xcolor}         

\usepackage{booktabs}
\usepackage{multirow}
\usepackage{amsmath}

\usepackage{pgfplots}
\usepackage{caption}
\usepackage{multicol}
\usepackage{enumitem}
\usepackage{arydshln}
\usepackage{tikz}
\usepackage{makecell}

\usepackage{subfig}
\usepackage{tikz}
\usepackage{pgfplots}
\usepackage{subcaption}

\workshoptitle{Efficient Reasoning}

\title{When Do Symbolic Solvers Enhance Reasoning in Large Language Models? }

\author{%
  Zhiyuan He \\
  University College London\\
  \texttt{nick.he.21@ucl.ac.uk} \\
  \And
  Dingmin Wang \\
  University of Oxford \\
  \texttt{dingmin.wang@cs.ox.ac.uk} \\
}

\begin{document}

\maketitle

\begin{abstract}

Large Reasoning Models (LRMs) achieve strong performance on complex reasoning tasks by generating long Chains of Thought (CoTs). However, this paradigm might incur substantial token overhead, especially when models ``overthink'' by producing lengthy reasoning chains, which can even lead to incorrect answers. A promising direction is the symbolic-solver-integrated approach, which leverages the code generation capabilities of LLMs to translate reasoning tasks into executable code and then solve them with a symbolic solver. In this paper, we explore an open question of \textit{when the conventional long-CoT can be enhanced by symbolic solvers}. Our experimental results show that the symbolic-solver-integrated method only helps when the problem requires limited implicit reasoning but involves an ample search space. The latest LLMs, like GPT-4o, show better performance on deductive problems with shallow reasoning depth, while the symbolic-solver-integrated method significantly improves the LLMs’ performance in constraint satisfaction problems that require repeated backtracks. When a declarative exemplar is provided, even CodeLlama-13B can outperform GPT-4o in difficult Zebra puzzles.

\end{abstract}

\section{Introduction}

Recent Large Language Models (LLMs) \citep{dubey2024llama, achiam2023gpt} have demonstrated outstanding capabilities in various domains including language understanding and generation tasks \citep{chang2024survey}. Chain-of-thought (CoT) prompting \citep{wei2023chainofthoughtpromptingelicitsreasoning} enables LLMs to articulate a step-by-step process to solve a problem or reach a conclusion. However, it is still an open problem whether language models can reason logically in the same way as humans do. In fact, recent works \citep{hassid2025don, kumar2025overthink} show that when models ``overthink'' by generating unnecessarily long reasoning chains, their accuracy can actually degrade, leading to incorrect answers. LLMs continue to struggle with complex, multi-step reasoning tasks \citep{huang2022towards, dziri2024faith}, likely due to the transformer \citep{NIPS2017_3f5ee243}’s inherent pattern-matching mechanism \citep{dziri2024faith, hu2024case}. This limitation significantly hampers the practical application of LLMs in fields such as education, law, and industrial code generation, where precise and reliable reasoning is critical \citep{venkit2024confidently, dahl2024hallucinating}.

To address this challenge, one research direction is \textit{symbolic-solver-integrated} methods that use external symbolic solvers to enhance the reasoning capabilities of language models \citep{giadikiaroglou2024puzzle}. These methods use LLMs' ability of code generation to translate the problems into formal language formulations, which can be interpreted by symbolic tools \citep{pan2023logic, ye2024satlm, gao2023pal, chen2022program}. 
However, these methods typically experiment on datasets comprising simple synthetic sentences that are easily translated into formal languages or involve a shallow reasoning depth.
Also, with the latest models like GPT-4o, benchmarks for many popular reasoning tasks have been raised. Do symbolic-solver-integrated methods still lead in performance? If not, for what types of problems should they be used?\par

We investigate these critical problems in integrating LLMs with symbolic solvers. We study how to model a problem efficiently and the advantage of symbolic-solver-integrated methods over CoT prompting. We experiment on five various datasets of three types: natural language arithmetic problems (GSM8K, GSM-Reversed, and GSMHard), complex constraint satisfaction problem ZebraLogic, and logic deduction problem EntailmentBank (task 2). We propose using Prolog and Python with symbolic libraries to improve the expressiveness of logical formulations, mainly due to the advanced data structures supported and the pretraining of LLMs on these programming languages. We also design exemplars to guide the language model, generating code in a declarative style and using concise statements that accurately capture the information in the problem.

Through the empirical analysis of the experiments, we find that it is crucial to model problems concisely into logical formulations for symbolic-solver-integrated methods. Our proposed prompting effectively increases the correctness rate of the generated code. With this prompting, the symbolic-solver-integrated methods enhance the reasoning ability in LLMs on constraint satisfaction problems where repeated backtracking is required. However, CoT reasoning is still advantageous in problems involving shallow deduction depths and implicit reasoning.

\section{Related Works}
\paragraph{Analysis of LLMs' reasoning ability} As transformer-based language models become more powerful and widely used, more research is increasingly focusing on the limits of these models in handling complex compositional tasks. \citet{zhou2023algorithms} indicate that the LLMs struggle with out-of-distribution generalization for algorithm tasks beyond a certain complexity. \citet{dziri2024faith} study the fundamental mechanism of LLMs solving reasoning tasks and observe that transformers generate the solutions relying on linear path matching or shortcut learning \citep{geirhos2020shortcut}. \citet{hu2024case} propose a congruent argument of LLMs using case-based reasoning instead of rule-based like humans. \citet{olausson2023linc} compared the performances and analyzed the errors of First Order Logic (FOL) formalization and CoT methods on ProofWriter and FOLIO datasets. 

\paragraph{Symbolic-solver-integrated methods for LLMs} This type of method involves an \textit{autoformalisation} \citep{wu2022autoformalization} stage where the problem in natural language is translated into expressions that external tools can interpret. \citet{gao2023pal} propose Program-aided Language Models (PAL) to translate problems into Python programs and solve them using a Python interpreter. \citet{chen2022program} use a similar methodology, Program of Thoughts (PoT), while it applies a different prompt that helps outperform PAL. More recently,  \citet{numina2024} won the first AIMO Progress Prize with their finetuned DeepSeekMath-Base 7B model by integrating the Python REPL. 
Another group of methods intends to translate the problems into FOL-based logical expressions to tackle the LLMs’ weakness in rule-based reasoning. \citet{olausson2023linc} translate the natural language premises and desired conclusion into FOL expressions. A symbolic FOL theorem prover will algorithmically determine the truth value of the conclusion. \citet{pan2023logic} experiment on a larger range of formulations, including Logical Programming, Constraint Satisfaction, and Boolean Satisfaction. \citet{ye2024satlm} design prompts to generate declarative specifications that are closer to the problem descriptions. These methods show promising results, especially on synthetic logical deduction datasets like FOLIO \citep{han2022folio}, ProofWriter \citep{tafjord2020proofwriter}, and ProntoQA \cite{saparov2022prontoqa}. 

In this work, we aim to explore the capability of symbolic-solver-integrated methods on problems that are more naturally stated and require richer semantic understanding. Instead of FOL-based expressions, we propose to model problems using more sophisticated programming languages like Prolog and Python, with the support of libraries proficient to different types of problems. We also design the prompt to elicit the LLMs to generate declarative codes in a sentence-to-sentence manner.

\section{Methodologies}
\subsection{Deductive Reasoning Problems}
We selected our datasets to ensure coverage of diverse reasoning types, allow control over reasoning difficulty, and include non-synthetic, natural language reasoning problems.

Compared to other types of reasoning (such as abductive reasoning, inductive reasoning, and analogical reasoning), deductive reasoning is more fundamental. 
We address three types of deductive problems: arithmetic reasoning, entailment reasoning, and constraint satisfaction logic reasoning.\par
Arithmetic reasoning test sets exhibit unique characteristics that impact the performance of different methods:
\begin{itemize}
    \item \textbf{Arithmetic calculations}: Arithmetic operations are basic, but they can become complex with large numbers.
    \item \textbf{Shallow reasoning depth}: Typically, these problems involve not deeply nested logical steps but relatively shallow deduction.
    \item \textbf{Implicit rules from word semantics}: The rules governing the problems are often hidden within the natural language semantics, making it challenging to formalize them into explicit mathematical or logical statements. 
\end{itemize}\par

Entailment is a fundamental concept in logic. It describes the relationship between statements that hold true when one statement logically follows from one or more statements. A logical argument is valid if the conclusion is entailed by the premises. That means that the conclusion is the consequence of the premises.\par

Constraint satisfaction logic reasoning is related to logic problems that can be suitably formulated as Constraint Satisfaction Problems (CSP). 
\citet{russell2016artificial} define a CSP as a triple \((\mathcal{X}, \mathcal{D}, \mathcal{C})\), where \(\mathcal{X}\) is a set of variables \(x_1, x_2, \ldots, x_n\); \(\mathcal{D}\) is a set of nonempty domains \(D_1, D_2, \ldots, D_n\) and each variable \(x_i\) has a nonempty domain \(D_i\) of possible values; \(\mathcal{C}\) is a set of constraints \(c_1, c_2, \ldots, c_m\) and each constraint \(c_i\) involves some subset of the variables and specifies the allowable combinations of values for that subset. An \textit{assignment} within a problem is represented by assigning specific values to a subset or all of the variables, for instance, \(\{X_i = v_i, X_j = v_j, \ldots\}\). 
When every variable has been assigned a value, it is a \textit{complete assignment}. In the context of a CSP, a solution is a complete assignment that meets all the constraints.\par
One classic example of CSPs is the Zebra puzzle, also known as the Einstein puzzle. In this problem, the goal is to determine specific features of a set of houses based on a series of given constraints. Each house has a unique combination of features such as color, owner, and pet. The solution space for the Zebra puzzle is given by \( H! \times F \), where \( H \) denotes the number of houses, and \( F \) denotes the number of features associated with each house. The constraints in the puzzle guide the elimination of invalid permutations, thereby narrowing down the possible solutions to the unique permutation that satisfies all given constraints.\par
While symbolic solvers can support a backtracking mechanism for CSPs, Chain-of-Thought (CoT) reasoning in LLMs faces unique challenges. To solve CSPs efficiently, CoT must employ heuristics to determine which variable to consider next and use trial-and-error to eliminate impossible values for a variable. The model could be overwhelmed by the exponentially increasing search space.

 \begin{figure*}[!ht]
  \centering

    \includegraphics[scale=0.75]{figures/prompts-and-generations/gsm/prompt-gsm-hard-python-new.png}

  \caption{One-shot prompt for GSM-Hard using a declarative sentence-to-sentence translation exemplar of Python code.} \label{prompt-gsm-hard-python}
  \end{figure*}
  
\subsection{Prompting With Declarative Exemplars}
\textit{Declarative programming} is a style of building the structure and elements of computer programs that expresses the logic of computation. In contrast, \textit{imperative programming} describes the control flow of how a program operates step by step.  Declarative programming focuses on high-level descriptions of its expected results. Apart from one of the most prevalent declarative programming languages, Prolog, we also use Python with symbolic libraries like \textit{Constraint} or \textit{SymPy} to generate code in a declarative style.\par
In our preliminary experiments, we observe that if we directly prompt the LLM to generate Python or Prolog code to solve the problem, it tends to generate code in an imperative way.
To generate declarative style codes for better performance, we provide the LLM with a declarative exemplar in the prompt. This exemplar involves translating the problem into declarative codes in a sentence-to-sentence manner, where each natural language sentence is copied into a comment, followed by its programming language representation. Figure~\ref{prompt-gsm-hard-python} illustrates a one-shot prompt with Python code using the SymPy library on the GSM-Hard dataset. Other prompts are illustrated in the Appendix~\ref{Prompts-and-generations}.



  
Notably, in arithmetic reasoning problems, we often need to solve for a value within an infinite, continuous range of real numbers. For Python, this is achieved using symbolic solvers from libraries such as SymPy. For Prolog, we utilize the \textit{Constraint Logic Programming over Reals} (clpr) library \citep{holzbaur1995ofai}. These methods rely on the symbolic solvers to find the unknown values that satisfy the given mathematical equations.\par

More specifically, here are the main formats that we use for problem auto-formalisation:
\begin{itemize}
    \item \textbf{Prolog} is a declarative language designed for expressing logical statements and reasoning. It is suitable for tasks involving rule-based logic and symbolic reasoning.
    \item \textbf{Prolog with clpr library} extends the capabilities of the standard Prolog. By incorporating constraint logic programming for real numbers, it allows for complex problem-solving involving numerical constraints.
    \item \textbf{Python with Constraint library} combines the flexibility and readability of Python with powerful constraint satisfaction capabilities. This approach enables us to define and solve logical problems using constraints within the popular Python environment.
    \item \textbf{Python with SymPy library} applies symbolic mathematics to represent and manipulate mathematical expressions and declarative logic. This method is beneficial for problems requiring symbolic computation and algebraic manipulation.
\end{itemize}

\section{Experimental Settings}


\begin{table}[ht]
\centering
\begin{tabular}{lcc}
\toprule
\textbf{Dataset Name} & \textbf{Size} & \textbf{Language + Library} \\
\midrule
GSM8K           & 1418 & \multirow{3}{*}{Prolog+clpr; Python+SymPy} \\
GSM-Reversed  & 1358 & \\
GSM-Hard      & 1319 & \\
\midrule
EntailmentBank      & 300 & Prolog \\
\midrule
ZebraLogic      & 1000 & Prolog; Python+Constraint \\
\bottomrule
\end{tabular}
\caption{Dataset details and the corresponding programming languages used.}
\label{tab:datasets}
\end{table}

In this paper, we use the following five datasets for testing.

\paragraph{Arithmetic reasoning problems} For this category, we use GSM8K \citep{cobbe2021training} and two of its variants, GSM-Reversed \citep{gao2024meta} and GSM-Hard \citep{gao2023pal}. 
GSM-Reversed tests a language model's reversed reasoning ability \citep{yu2023metamath}. 
The final answer to the original GSM8K question is given in the prompt, but part of the input value is concealed and asked for a solution. GSM-Hard is constructed by replacing a number in the original GSM8K problem with a random larger number, up to seven digits. It tests the language model's mathematical abilities with large numbers, non-integer numbers, and negative numbers.

\paragraph{Entailment reasoning} We experiment on the EntailmentBank \citep{dalvi2021explaining} dataset, which is a rich collection of entailment trees designed to evaluate models in natural language understanding and reasoning. Each question requires to create a tree of multi-premise entailment steps from facts that are known, through intermediate conclusions, until the desired hypothesis is reached.

\paragraph{Constraint satisfaction reasoning problems} We evaluate CoT and symbolic-solver-integrated methods on ZebraLogic \citep{Lin_ZeroEval_A_Unified_2024}, a dataset designed to test on CSP and offers tunable difficulty levels, making it ideal for comprehensive evaluations. 
We extensively tested all available sizes of ZebraLogic puzzles, ranging from 2x2 to 6x6, to compare the performance of the two types of methods under different reasoning difficulties.\par
For each dataset, we compare the LLM's CoT reasoning against the symbolic-solver-integrated methods. We choose suitable logical formulations for each method to fit the question's purpose and formulate the question with higher accuracy. The details are presented in table \ref{tab:datasets}.\par

We experiment with one of the state-of-the-art LLMs for code and reasoning: GPT-4o. We also test with a smaller Python-specialized LLM \textit{CodeLlama-13b-Python-hf}\footnote{Throughout the rest of this paper, we use the abbreviations 'CodeLlama13B' or 'CodeL13B' to refer to the model \textit{CodeLlama-13b-Python-hf}.}. We use greedy decoding for reproducibility and to reduce randomness. We use zero-shot, or one-shot if the model is required to generate code in a certain template.

For the methods that use code generation, we use the self-refinement mechanism \citep{pan2023logic}, which returns the error message to the LLMs and asks for regeneration when the interpreter reports a syntax error. We set the maximum retries to be ten times for all experiments.

\section{Results}

\begin{table*}[ht]
\centering
\setlength{\tabcolsep}{3pt} 
\renewcommand{\arraystretch}{0.6}
\small
\begin{tabular}{ll|ccc|ccc|c}
\toprule
\multirow{2}{*}{\textbf{Model}} & 
\multirow{2}{*}{\textbf{Prompting}} & 
\multicolumn{3}{c}{\textbf{Arithmetic}} &  
\multicolumn{3}{c}{\textbf{ZebraLogic}} & 
\multirow{2}{*}{\textbf{Entailment}} 

\\
 & & \textbf{GSM8K} 
 & \textbf{GSM-Rev} 
 & \textbf{GSM-Hard} 
 & \textbf{easy} 
 & \textbf{hard} 
 & \textbf{average} 
 &  \\
\midrule

\textbf{GPT-4o} & \textbf{CoT} & 94.6 & \textbf{90.3} & 69.1 
& 77.9\(^\ast\) & 8.9\(^\ast\) & 28.2\(^\ast\)
& \textbf{81.8} \\
\textbf{Claude 3.5 Sonnet} & \textbf{CoT} & 95.6  & - & - 
& 87.5\(^\ast\) & 12.4\(^\ast\) & 33.4\(^\ast\)
& - \\
\textbf{Llama-3.1-405B} & \textbf{CoT} & \textbf{95.9} & - & -
& 87.1\(^\ast\) & 11.4\(^\ast\) & 32.6\(^\ast\)
& - \\
\textbf{code-davinci-002} & \textbf{PAL} & 72.0 & - & 61.2 
& - & - & -
& - \\
\textbf{code-davinci-002} & \textbf{PoT} & 71.6 & - & 61.8 
& - & - & -
& - \\
\textbf{CodeLlama13B} & \textbf{CoT} & 22.1 & - & - 
& - & - & -
& - \\
\midrule
\textbf{GPT-4o} & \textbf{PL-ZS} & 85.8 & 76.2 & 71.3 
& 64.6 & 15.0 & 28.9
& 27.4 \\
\textbf{GPT-4o}& \textbf{decl. PL-1S} & 92.6 & 87.8 & \textbf{78.0} 
& \textbf{96.8} & 44.0 & 58.8
& - \\
\textbf{GPT-4o} & \textbf{PY-ZS} & 87.7 & 87.6 & 75.7 
& 69.7 & 26.5 & 38.1
& - \\
\textbf{GPT-4o} & \textbf{decl. PY-1S} & 90.1 & 87.5 & 72.6 
& 96.4 & \textbf{62.1} & \textbf{71.7}
& - \\
\textbf{CodeLlama13B} & \textbf{decl. PL-1S} & 43.0 & 34.1 & 35.8 
& 62.5 & 14.6 & 28.5
& - \\
\textbf{CodeLlama13B} & \textbf{decl. PY-1S} & 38.9 & 29.1 & 34.7 
& 74.3 & 20.0 & 35.2
& - \\
\bottomrule
\end{tabular}
\caption{Performance on five datasets with different methods. The prompting methodologies are written in the following abbreviations: decl. (declarative style exemplar), 1S (one-shot), ZS (zero-shot), PL (Prolog), and PY (Python). For example, \textit{decl. PL-1S} indicates that the LLM is prompted to generate Prolog code using one exemplar written in a declarative style.}
\label{main-results}
\end{table*}

Experimental results in Table~\ref{main-results} show that GPT-4o CoT gets an accuracy greater than 90\% for GSM8K and GSM-Reversed and outperforms the methods using programming languages\footnote{There is concern that LLMs perform well due to dataset contamination. In \citet{zhang2024careful}, GPT-4 performs better on a new test set GSM1K than GSM8K. This result validates its genuine reasoning capability.}. CoT reasoning demonstrates its ability to handle shallow deductive reasoning effectively and to interpret and process the natural language semantics of the problem. These abilities allow it to navigate the implicit rules and perform the necessary calculations.
In contrast, symbolic-solver-integrated methods may fail to capture the implicit rules in certain problems, as demonstrated in the failure modes discussed in Section~\ref{failure-modes}.
\par

The smaller LLM shows a different result. For CodeLlama13B, the methods using programming languages significantly outperform those using COT. Even their performance on GSM-Hard is better than COT's performance on the original GSM8K. It shows that symbolic-solver-integrated methods could be advantageous when the problem involves a reasoning complexity beyond the LLM's capability.

For GSM-Hard and ZebraLogic, the symbolic-solver-integrated methods with one-shot prompts significantly outperform CoT. Claude 3.5 Sonnet and Llama-3.1-405B are the top two models in the leaderboard of ZebraLogic\footnote{\url{https://huggingface.co/spaces/allenai/ZebraLogic}}. GPT-4o with Prolog or Python using one-shot prompts outperforms them in all metrics: \textit{easy}, \textit{hard}, and \textit{average}.\par
The results on GSM-Hard show that LLMs encounter significant challenges with large numbers.
This is where symbolic solvers have a distinct advantage. These methods excel at performing precise calculations using mathematical operations and logic, regardless of the size of the numbers.\par
LLMs with CoT also struggle with complex CSPs. They outperform zero-shot symbolic-solver-integrated methods only on simple problems. We observe that the LLM tends to use rule-based deductions rather than a trial-and-error strategy on CSPs. As the problem size increases, the interrelationships between features become more intricate. It becomes difficult to solve through the token-by-token autoregressive generation. Frequently, unfaithful reasoning is observed, especially in the hard-level puzzles.

Symbolic formalization, on the other hand, systematically translates each natural language sentence into a corresponding mathematical equation or logical statement. This method relies on symbolic solvers (e.g., SymPy library in Python and clpr library in Prolog) to interpret the translated formal language program and find solutions that satisfy all constraints. Symbolic solvers can reason robustly and faithfully. They also employ a backtracking mechanism suitable for problems requiring extensive search trees. When a variable has multiple potential values, one should attempt a value and backtrack if it leads to conflicts. This backtracking allows the solver to explore multiple potential solutions and revert to previous states if a conflict arises.\par


The one-shot prompting for symbolic-solver-integrated methods outperforms their corresponding zero-shot prompting over all the experiments except for solving the Reversed and Hard variations of GSM8K using Python. Prompting LLMs to generate codes in declarative style effectively improves the quality of generated code.\par

In zero-shot cases, Prolog code generation involves considerably more syntax errors than Python. This result is probably due to Python's widespread use and extensive repositories that provide ample training data.\par

GPT-4o with Prolog gets much lower performance in the EntailmentBank dataset than GPT-4o-CoT. 
Modeling more natural and sophisticated problems with programs is significantly challenging.
This difficulty arises from several inherent complexities in natural language (illustrated in Section~\ref{failure-modes}) that are hard to be captured into formal languages. 


 \begin{figure}[!ht]
  \centering
  
  \subfloat[An implicit semantic rule omitted in EntailmentBank]{
  \includegraphics[scale=0.8]{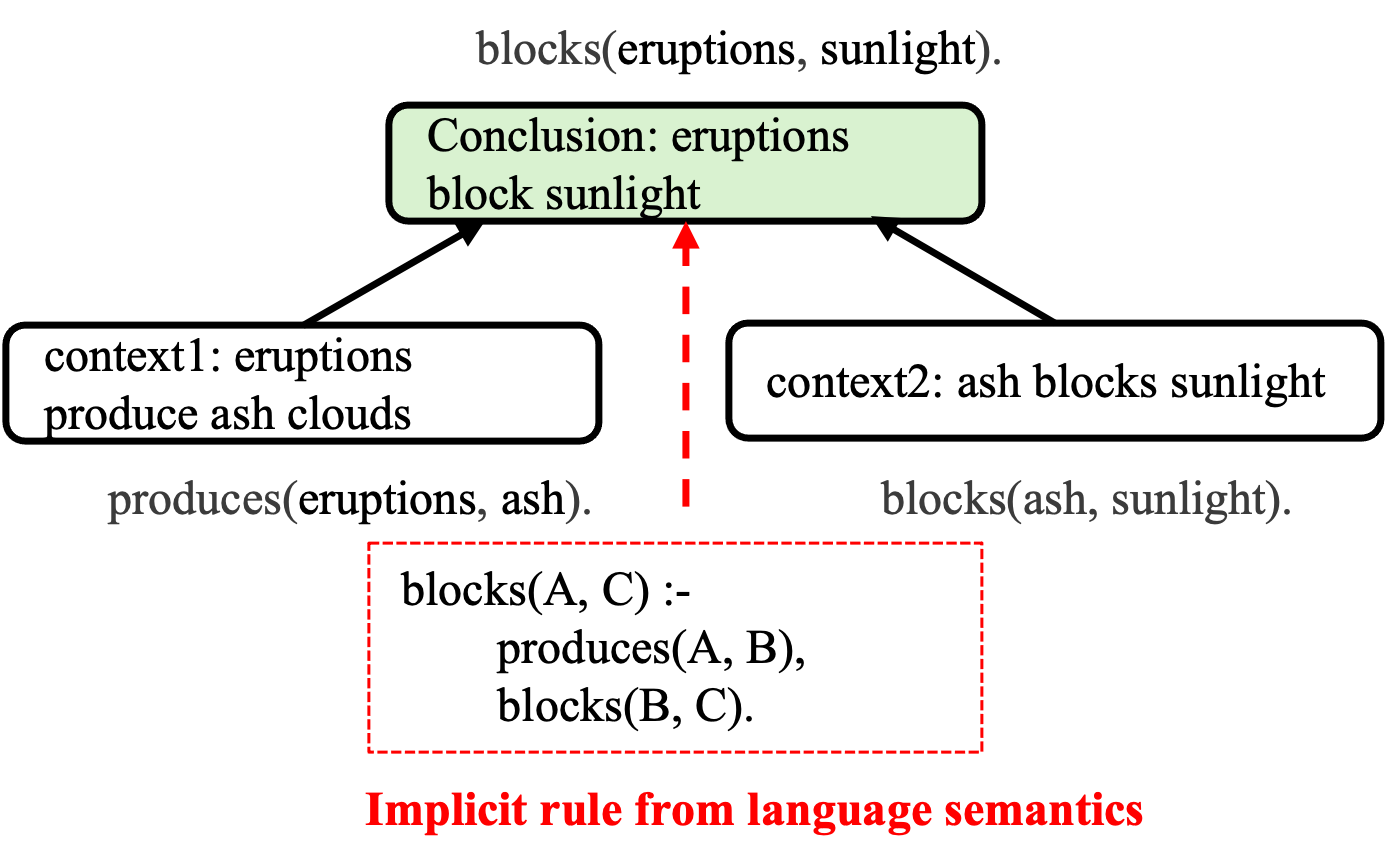}
 \label{entailment-implicit-rule}
  }

  \subfloat[Incorrect implicit reasoning in GSM8K-reversed]{
  \includegraphics[scale=0.8]{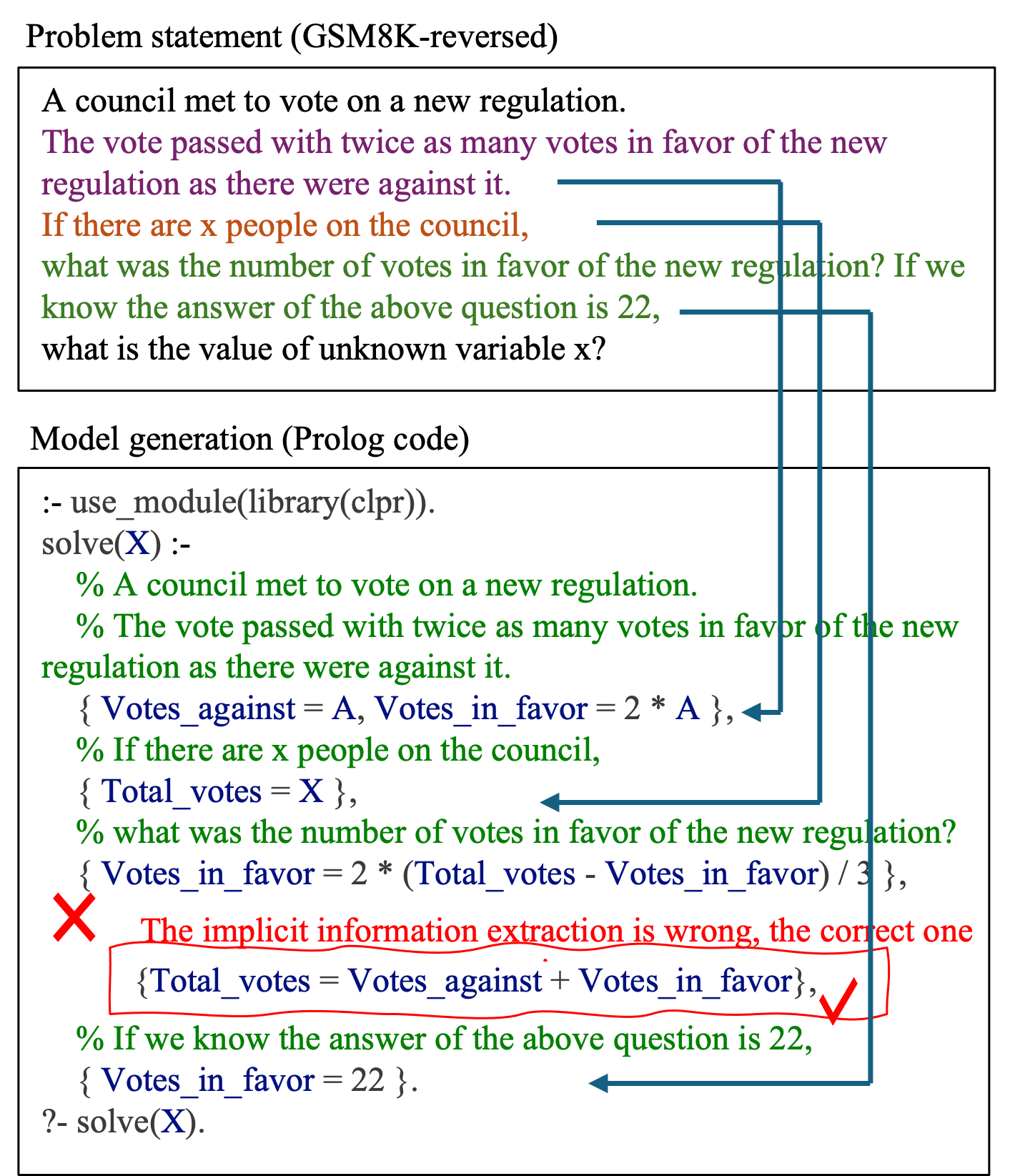}
 \label{gsm-implicit-info}
  }

  \caption{Examples of LLMs failing to reason about implicit information in code generation}
  \label{implicit-rules}
  \end{figure}
  
\section{Analysis}
\subsection{Ablation Tests}

\paragraph{One-shot vs Zero-shot} Table~\ref{main-results} shows that one-shot prompting generally results in better performance. The improvement is more outstanding on the ZebraLogic dataset. The only exception is in GSM-Reversed and GSM-Hard with Python, where the accuracy decreases marginally.

\paragraph{Retries in code execution} 
Furthermore, we prompt the LLMs to regenerate code when a syntax error is detected. Table~\ref{Code-Execution} illustrates the impact on performance when the maximum number of attempts is set to one, two, and ten. Overall, the retry mechanism is particularly beneficial for Prolog code generation and for generating code with smaller models, whereas the larger model can generate code with less syntax error at the first attempt, especially with Python.



\begin{table}[htbp]
\centering
\begin{tabular}{c|c|ccc|ccc}
\toprule
\textbf{Model} & \textbf{Prompt} 
& \multicolumn{3}{c}{\textbf{GSM-Hard}} 
& \multicolumn{3}{c}{\textbf{ZebraLogic}} \\
 & 
 & \textbf{1} & \textbf{2} & \textbf{10} & \textbf{1} & \textbf{2} & \textbf{10}\\
\midrule
\textbf{GPT-4o} & PL-ZS & 49 & 63 & 71 & 43 & 43 & 43 \\
\textbf{GPT-4o} & PL-1S & 73 & 77 & 78 & 83 & 83 & 83 \\
\textbf{GPT-4o} & PY-ZS & 75 & 76 & 76 & 49 & 51 & 51 \\
\textbf{GPT-4o} & PY-1S & 71 & 73 & 73 & 84 & 84 & 84 \\
\textbf{CodeL13B} & PL-1S & 33 & 35 & 36 & 36 & 36 & 36 \\
\textbf{CodeL13B} & PY-1S & 29 & 33 & 35 & 32 & 35 & 41 \\

\bottomrule
\end{tabular}
\caption{Accuracy in percentage with 1, 2, 10 of maximum number of attempts allowed. For ZebraLogic, we count the 160 test cases in 4 middle-level sizes: 3*3, 3*4, 4*3 and 4*4.}
\label{Code-Execution}
\end{table}

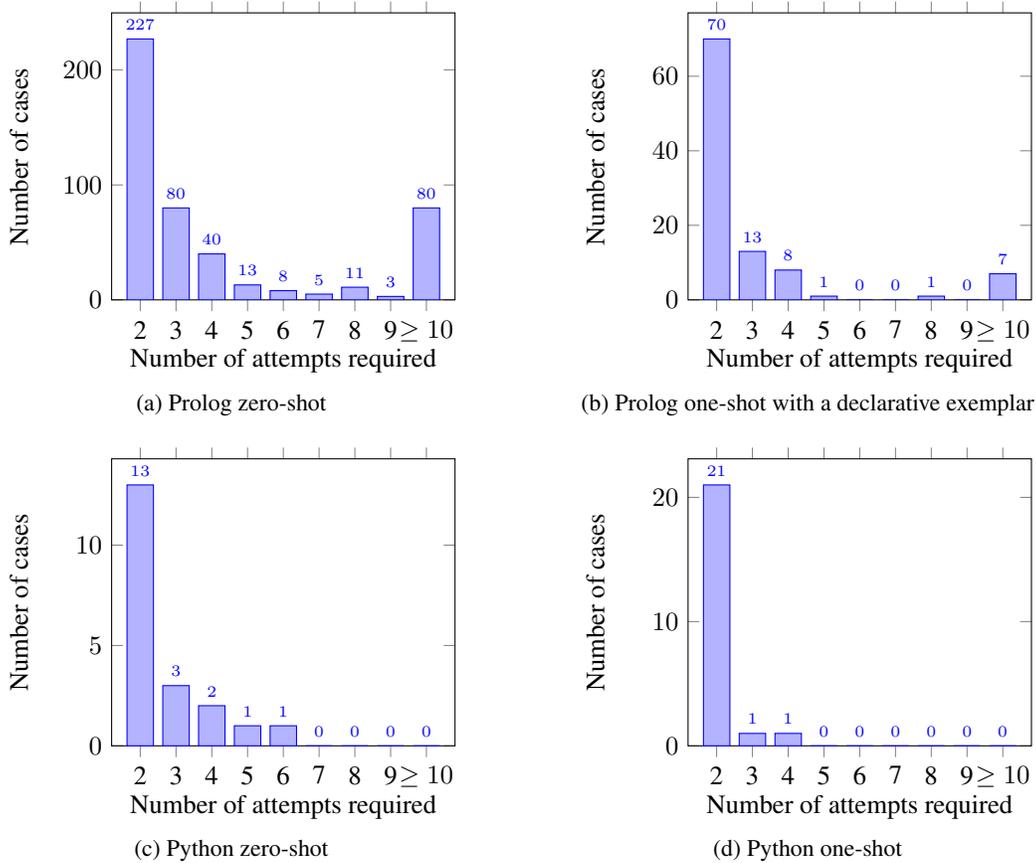
\begin{figure*}[t!]
    \centering
    
    \subfloat[Prolog zero-shot]{
        \begin{tikzpicture}
            \begin{axis}[
                height=5.4cm, 
                width=0.44\linewidth, 
                ybar,
                ymin=0,
                bar width=10pt,
                enlarge x limits=0.1,
                xlabel={Number of attempts required},
                ylabel={Number of cases},
                xtick={1,2,3,4,5,6,7,8,9},
                xticklabels={2,3,4,5,6,7,8,9,{\(\geq 10\)}},
                nodes near coords,
                every node near coord/.append style={font=\tiny},
            ]
                \addplot coordinates {(1,227) (2,80) (3,40) (4,13) (5,8) (6,5) (7,11) (8,3) (9,80)};
            \end{axis}
        \end{tikzpicture}
        \label{gsm-hard-pl-0s}
    }
    \hfill
    \subfloat[Prolog one-shot with a declarative exemplar]{
        \begin{tikzpicture}
            \begin{axis}[
                height=5.4cm, 
                width=0.44\linewidth, 
                ybar,
                ymin=0,
                bar width=10pt,
                enlarge x limits=0.1,
                xlabel={Number of attempts required},
                ylabel={Number of cases},
                xtick={1,2,3,4,5,6,7,8,9},
                xticklabels={2,3,4,5,6,7,8,9,{\(\geq 10\)}},
                nodes near coords,
                every node near coord/.append style={font=\tiny},
            ]
                \addplot coordinates {(1,70) (2,13) (3,8) (4,1) (5,0) (6,0) (7,1) (8,0) (9,7)};
            \end{axis}
        \end{tikzpicture}
        \label{gsm-hard-pl-1s}
    }


    \subfloat[Python zero-shot]{
        \begin{tikzpicture}
            \begin{axis}[
                height=5.4cm, 
                width=0.44\linewidth, 
                ybar,
                ymin=0,
                bar width=10pt,
                enlarge x limits=0.1,
                xlabel={Number of attempts required},
                ylabel={Number of cases},
                xtick={1,2,3,4,5,6,7,8,9},
                xticklabels={2,3,4,5,6,7,8,9,{\(\geq 10\)}},
                nodes near coords,
                every node near coord/.append style={font=\tiny},
            ]
                \addplot coordinates {(1,13) (2,3) (3,2) (4,1) (5,1) (6,0) (7,0) (8,0) (9,0)};
            \end{axis}
        \end{tikzpicture}
        \label{gsm-hard-py-0s}
    }
    \hfill
    \subfloat[Python one-shot]{
        \begin{tikzpicture}
            \begin{axis}[
                height=5.4cm, 
                width=0.44\linewidth, 
                ybar,
                ymin=0,
                bar width=10pt,
                enlarge x limits=0.1,
                xlabel={Number of attempts required},
                ylabel={Number of cases},
                xtick={1,2,3,4,5,6,7,8,9},
                xticklabels={2,3,4,5,6,7,8,9,{\(\geq 10\)}},
                nodes near coords,
                every node near coord/.append style={font=\tiny},
            ]
                \addplot coordinates {(1,21) (2,1) (3,1) (4,0) (5,0) (6,0) (7,0) (8,0) (9,0)};
            \end{axis}
        \end{tikzpicture}
        \label{gsm-hard-py-1s}
    }

    \caption{Investigate the distribution of the number of retries needed to recover from syntax errors of the generated Prolog and Python code. We use GSM-Hard dataset, which contains 1319 problems in total, as an example.}
    \label{retries-gsm-hard-gpt4o}
\end{figure*}

Figure~\ref{retries-gsm-hard-gpt4o} illustrates the distribution of retries required to obtain executable code using GPT-4o across Prolog and Python settings.
For Prolog (Figures~\ref{gsm-hard-pl-0s} and \ref{gsm-hard-pl-1s}), the zero-shot setting shows that many cases require several retries, with 80 out of 1319 problems still failing to produce executable code after ten attempts. Using a one-shot prompt significantly improves the success rate, reducing the number of cases that need multiple retries.
For Python (Figures~\ref{gsm-hard-py-0s} and \ref{gsm-hard-py-1s}), the zero-shot setting already yields high-quality executable code with few retries, and the one-shot prompt further improves the success rate by reducing the number of cases that require multiple attempts.
\par
We also investigate the distribution of retries for ZebraLogic using GPT-4o with Prolog/Python codes, as shown in Appendix~\ref{Histogram Of Retries}. The results are consistent with GSM-Hard, and the same appendix includes distributions for both GSM-Hard and ZebraLogic using CodeLlama13B with Prolog/Python codes and one-shot prompts, indicating that CodeLlama13B struggles more to generate executable code than GPT-4o. This result is not surprising because of the significant difference in the model size.

\subsection{Declarative Exemplar Prompting}

Prompting with a declarative exemplar enhances the correctness rate under almost all the experiment settings. Our proposed method addresses and resolves two types of problems shown in the zero-shot experiments.

\paragraph{LLMs tend to generate problematic imperative codes}
We observe that when directly prompting the LLM to generate Python or Prolog code to solve the problem, it tends to generate code in an imperative way. It often skips intermediate reasoning steps and leads to incorrect results. In contrast, with declarative prompting, the symbolic solver faithfully handles all intermediate reasoning rather than relying on the LLMs. The examples of comparing the two methods are illustrated in Appendix~\ref{appendix-failure-modes}. \par

\paragraph{Generated Prolog codes have excessive predicates} LLMs tend to convert concepts into predicates in Prolog codes, generating excessive and unnecessary predicates. This result complicates the code and leads to more errors. We speculate that this issue arises because the LLM is pre-trained on Prolog code generation using logic deduction problems, where concepts are normally expressed using predicates. Our declarative exemplars help to translate the problem into concise code with a clean structure, significantly improving correctness.  Appendix~\ref{appendix-failure-modes} illustrates this failure mode and the subsequent improvement after applying the exemplar.\par

\begin{table*}[!ht]
\centering
\subfloat[GPT-4o with Python code (one-shot)]{
    \begin{tabular}{c|ccccc}
    \toprule
    \multirow{2}{*}{\textbf{Houses}} & \multicolumn{5}{c}{\textbf{Number of Features}} \\
     & \textbf{2} & \textbf{3} & \textbf{4} & \textbf{5} & \textbf{6} \\
    \midrule
    \textbf{2} & 100 & 100 & 100 & 97.5 & 97.5 \\
    \textbf{3} & 92.5 & 87.5  & 90 & 87.5 & 72.5 \\
    \textbf{4} & 100 & 75 & 75 & 75 & 65 \\
    \textbf{5} & 90 & 70 & 62.5 & 42.5 & 30 \\
    \textbf{6} & 87.5 & 50 & 17.5 & 15.0 & 12.5 \\
    \bottomrule
    \end{tabular}
    \label{tab:zebralogic-gpt-4o-python}
}
\subfloat[GPT-4o with Prolog code (one-shot)]{
    \begin{tabular}{c|ccccc}
    \toprule
    \multirow{2}{*}{\textbf{Houses}} & \multicolumn{5}{c}{\textbf{Number of Features}} \\
     & \textbf{2} & \textbf{3} & \textbf{4} & \textbf{5} & \textbf{6} \\
    \midrule
    \textbf{2} & 100 & 100 & 100 & 97.5 & 97.5 \\
    \textbf{3} & 92.5 & 90.0  & 90.0 & 77.5 & 65.0 \\
    \textbf{4} & 97.5 & 80.0 & 72.5 & 70.0 & \textbf{0} \\
    \textbf{5} & 85.0 & 75 & \textbf{0} & \textbf{0} & \textbf{0} \\
    \textbf{6} & 80.0 & \textbf{0} & \textbf{0} & \textbf{0} & \textbf{0} \\
    \bottomrule
    \end{tabular}
    \label{tab:zebralogic-gpt-4o-prolog}
}\\
\subfloat[CodeLlama13B with Python code (one-shot)]{
    \begin{tabular}{c|ccccc}
    \toprule
    \multirow{2}{*}{\textbf{Houses}} & \multicolumn{5}{c}{\textbf{Number of Features}} \\
     & \textbf{2} & \textbf{3} & \textbf{4} & \textbf{5} & \textbf{6} \\
    \midrule
    \textbf{2} & 97.5 & 85.0 & 75.0 & 67.5 & 77.5 \\
    \textbf{3} & 57.5 & 60.0  & 42.5 & 40.0 & 25.0 \\
    \textbf{4} & 40 & 40 & 22.5 & 25 & 15 \\
    \textbf{5} & 40 & 22.5 & 7.5 & 5 & 0.15 \\
    \textbf{6} & 7.5 & 12.5 & 0 & 0 & 0 \\
    \bottomrule
    \end{tabular}
    \label{tab:zebralogic-cl13b-python}
}
\subfloat[Claude-3.5-Sonnet with CoT]{
\begin{tabular}{c|ccccc}
    \toprule
    \multirow{2}{*}{\textbf{Houses}} & \multicolumn{5}{c}{\textbf{Number of Features}} \\
     & \textbf{2} & \textbf{3} & \textbf{4} & \textbf{5} & \textbf{6} \\
    \midrule
    \textbf{2} & 100 & 100 & 90 & 95 & 90 \\
    \textbf{3} & 75 & 62.5  & 37.5 & 30 & 17.5 \\
    \textbf{4} & 55 & 37.5 & 10 & 2.5 & 2.5 \\
    \textbf{5} & 15 & 10 & 0 & 0 & 0 \\
    \textbf{6} & 5 & 0 & 0 & 0 & 0 \\
    \bottomrule
    \end{tabular}
\label{tab:zebralogic-claude-cot}
}

\caption{Accuracy of ZebraLogic puzzles}
\label{tab:zebralogic-exps}
\end{table*}

\subsection{Difficulty Levels of Reasoning}
We investigate the impact of the difficulty levels of reasoning on the ZebraLogic dataset. 
The number of houses and the number of features define the size of a ZebraLogic puzzle ranging from 2*2 to 6*6. According to the results generated and the ground truth in the ZeroEval repository, we calculated the accuracy of each puzzle size for Claude-3-5-sonnetand GPT-4o. We compare these two baselines with the accuracies using Python and Prolog codes with GPT-4o and CodeLlama13B. Part of the above results are illustrated in Table~\ref{tab:zebralogic-exps}, and the rest are in the Appendix~\ref{Appendix-ZebraLogic-Results}.\par
Claude 3.5 holds the top position in the leaderboard. This model and GPT-4o perform well in easy puzzles, particularly in the first row of tables. However, the methods using code generation with GPT-4o still outperform them by accuracies larger than 97.5 in the first row.\par
When the size of the puzzle increases, the performance of Claude 3.5 and GPT-4o rapidly decreases. For example, the accuracy of sizes 3*3 and 4*4 from Claude 3.5 are 62.5 and 10, respectively. While the methods using code generation with GPT-4o achieve 90 and 72.5 respectively. 
Even CodeLlama13B outperforms the two CoT baselines on the hard-level puzzles, except for the size 4*2.
These results show that code generation with declarative methods is more advantageous on some complex CSPs like ZebraLogic.\par

\paragraph{Errors made by LLMs with CoT prompting in complex reasoning} We observe three representative types of errors that might lead to LLMs' failure.\footnote{Examples are illustrated in the Appendix~\ref{appendix-failure-modes}}
\begin{enumerate}[label=\textbullet., leftmargin=*, labelsep=0.5em]
    \item \textbf{Partial Deduction}: At a reasoning step, the LLM could deduce only one conclusion when multiple are possible. 
    \item \textbf{Unfaithful Deduction}: The reasoning process sometimes produces deductions that are not logically consistent with the given facts, leading to incorrect conclusions.
    \item \textbf{Misinterpretation of Problem Statements}: The LLM may misunderstand the problem statement, leading to incorrect formalization and flawed reasoning paths.
\end{enumerate}

  

\subsection{Failure Modes of Symbolic-Solver-Integrated Methods}
\label{failure-modes}

\paragraph{Formalizing implicit rules} Natural language is rich with implicit semantic rules that guide understanding and interpretation. These rules are often not explicitly stated but are understood through context and commonsense reasoning. Formalizing these implicit rules in formal languages is challenging because it requires making all underlying assumptions and contextual knowledge explicit. Figure~\ref{implicit-rules} demonstrates such failures on different datasets.

\paragraph{Formalizing complex sentence structures} Large Language Models often struggle with complex natural sentences, particularly those containing intricate clause structures. For example, \textit{``when carbon dioxide in the atmosphere is absorbed by plants, the amount of carbon dioxide in the atmosphere is reduced''} might lead to misinterpretation or failure to capture the nuanced relationships between different sentence parts like in conditional or dependent clauses.
\paragraph{Formalizing Modal Logic} Modal Logic \citep{russell2016artificial} deals with necessity and possibility, for example statements like \textit{It is necessary that...} or \textit{It is possible that...}. Although modal logics do not exceed the expressiveness of FOL, they need to be meticulously defined, such as using Kripke models \citep{Kripke1963SemanticalAO}. This process can pose huge challenge the code generation capabilities of a language model.

\section{Conclusion}
Large language models (LLMs) show good performance in shallow reasoning problems like GSM8K and its variants. Using the symbolic-solver-integrated method helps with the reasoning capability for less powerful language models, such as CodeLlama-13B, or when the problems are complex and require trial-and-error or repeated backtracks, such as hard Zebra puzzles. Our proposal to use one-shot prompting with a declarative exemplar translates the natural language description of the problem into a declarative code in a sentence-to-sentence manner. This method improves the quality of the code generated by LLMs in the symbolic-solver-integrated method, both using a traditional declarative language (Prolog) and a popular language (Python).

\bibliographystyle{unsrtnat}
\bibliography{references} 








\appendix

\appendix

\clearpage
\section{Prompts And Generations}
\label{Prompts-and-generations}

\begin{figure}[!ht]
  \centering
    \includegraphics[scale=1]{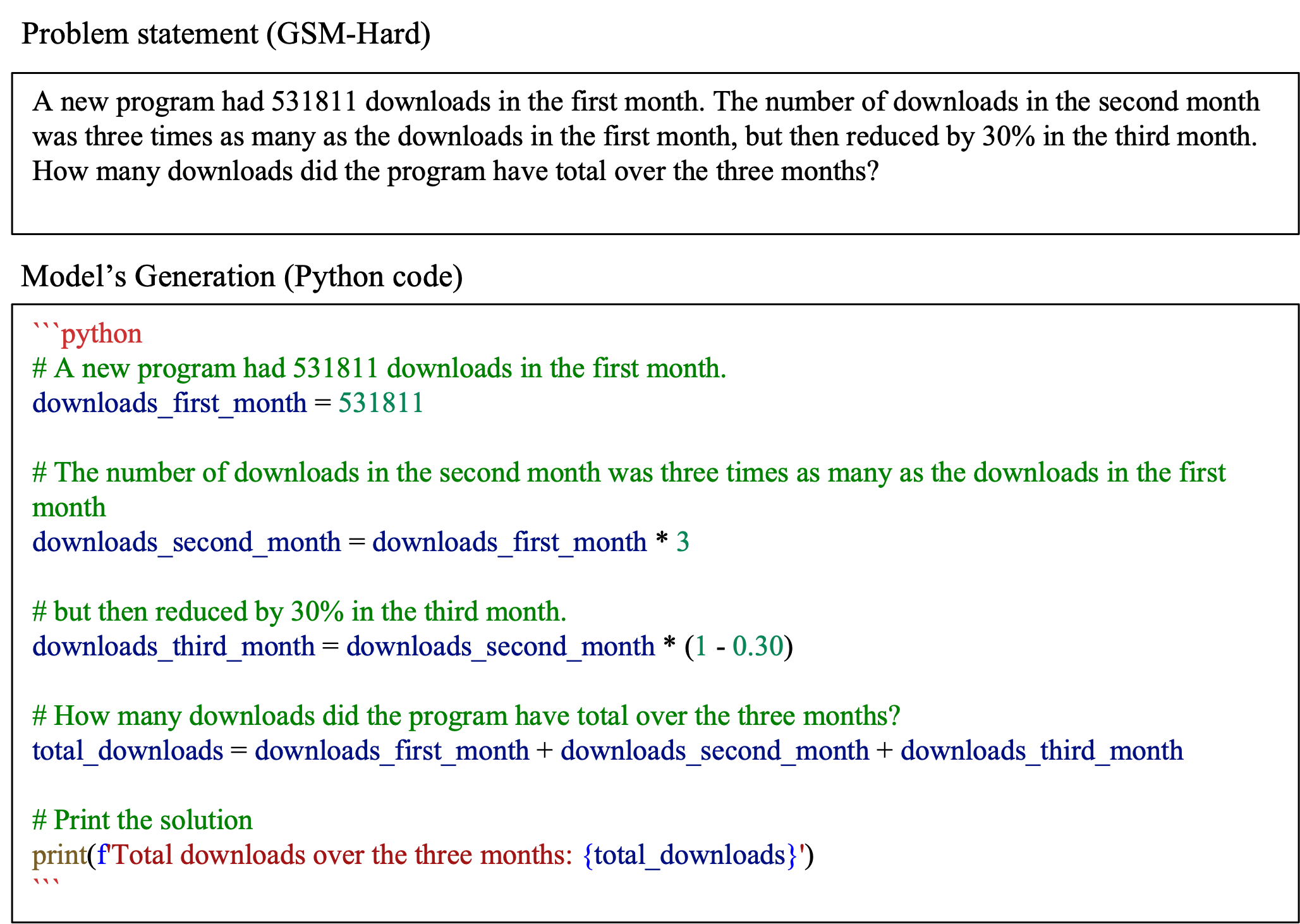}
  \caption{GPT-4o with Python on GSM-Hard} \label{generation-gsm-hard-python}
\end{figure}

   \begin{figure*}[!ht]
  \centering
    \includegraphics[scale=1]{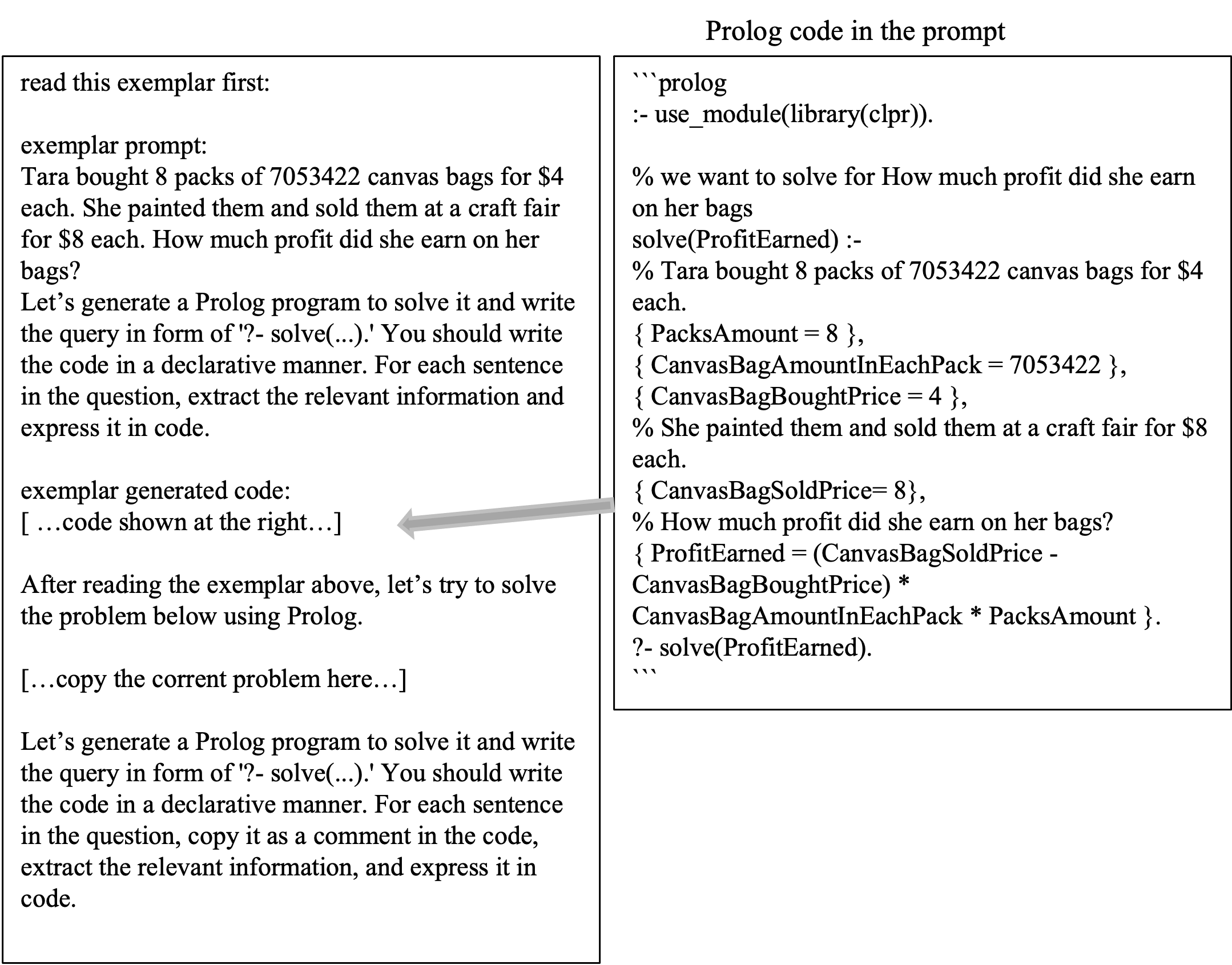}
  \caption{Prompt with a declarative Prolog exemplar for GSM-Hard} \label{prompt-gsm-hard-prolog}
  \end{figure*}

 \begin{figure*}[!ht]
  \centering
    \includegraphics[scale=1]{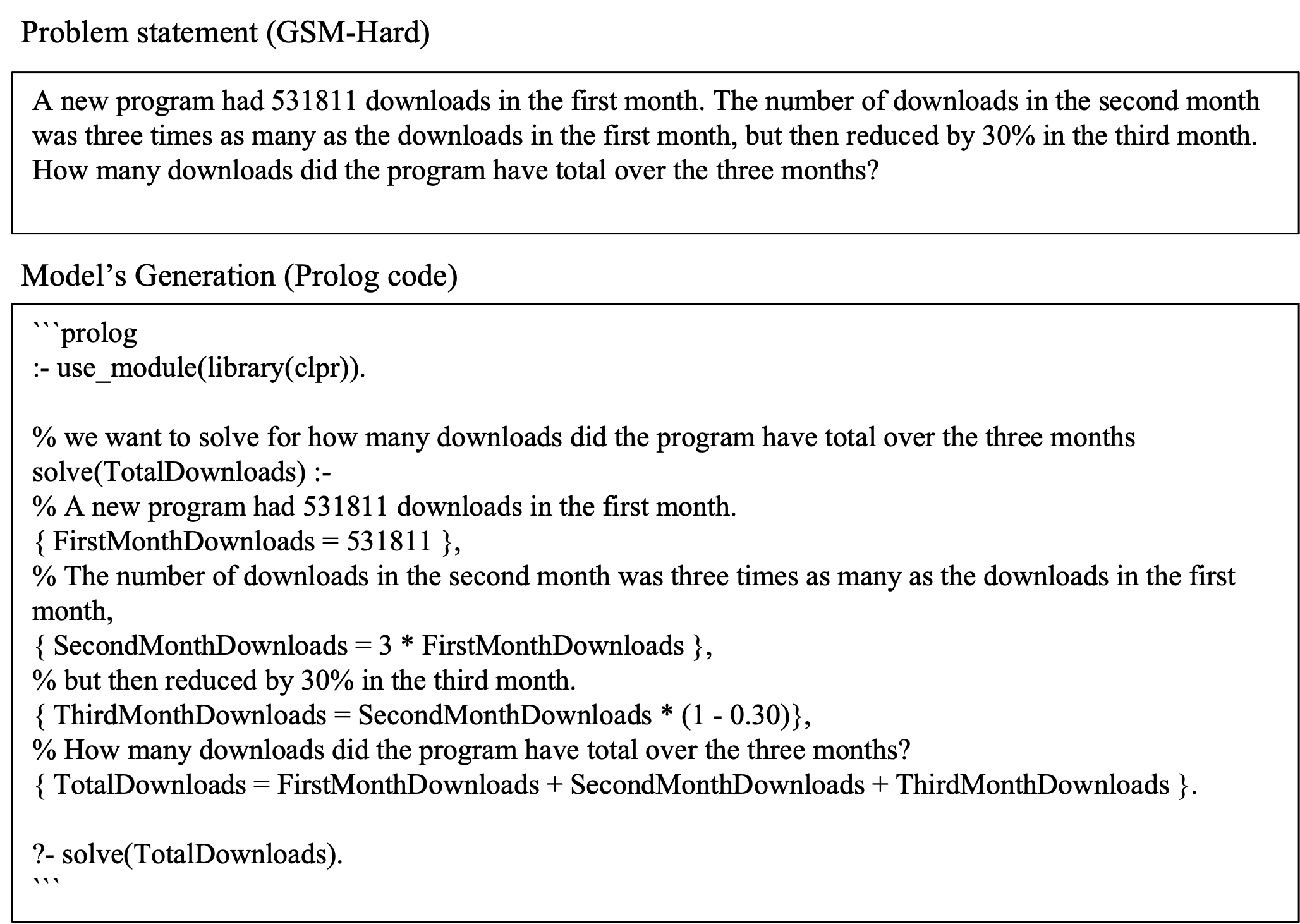}
  \caption{Generation from GPT-4o with Prolog on GSM-Hard} \label{generation-gsm-hard-prolog}
  \end{figure*}

   \begin{figure*}[!ht]
  \centering
    \includegraphics[scale=0.85]{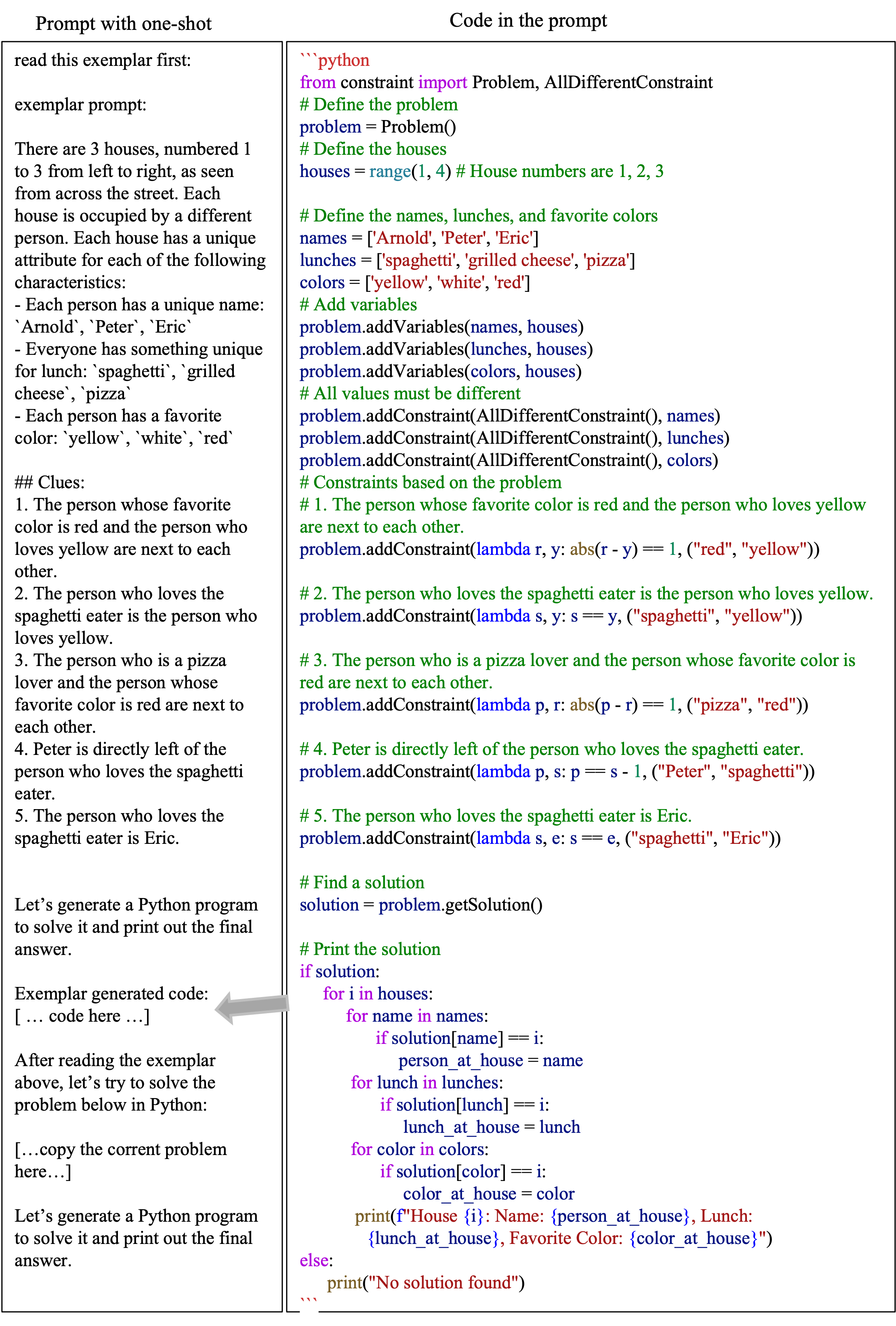}
  \caption{Prompt with a declarative Python exemplar for ZebraLogic} 
  \label{prompt-zebralogic-python}
  \end{figure*}

     \begin{figure*}[!ht]
  \centering
    \includegraphics[scale=0.82]{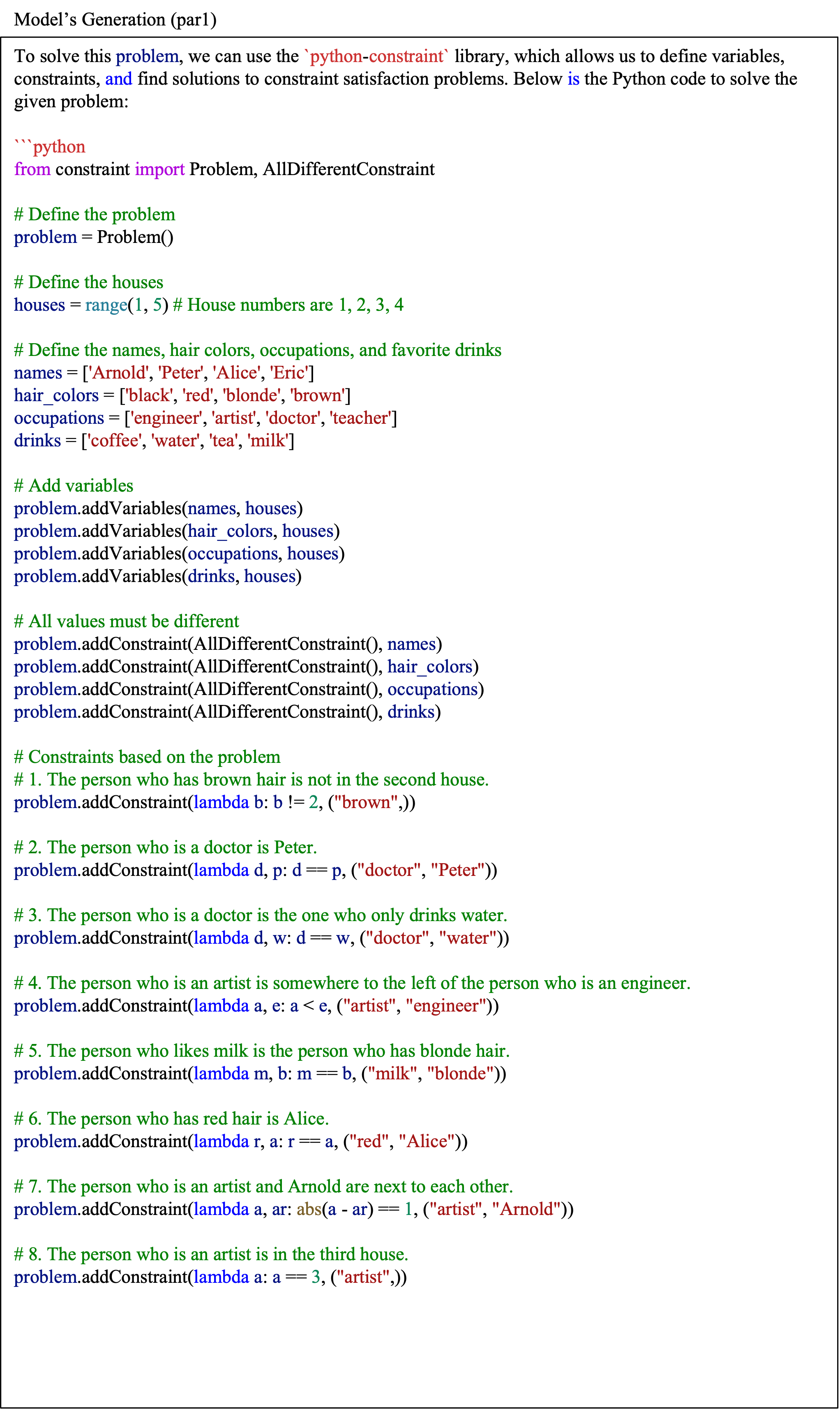}
  \caption{Generation from GPT-4o with Python on ZebraLogic (Part1)} 
  \label{eneration-zebralogic-python-part1}
  \end{figure*}

       \begin{figure*}[!ht]
  \centering
    \includegraphics[scale=1]{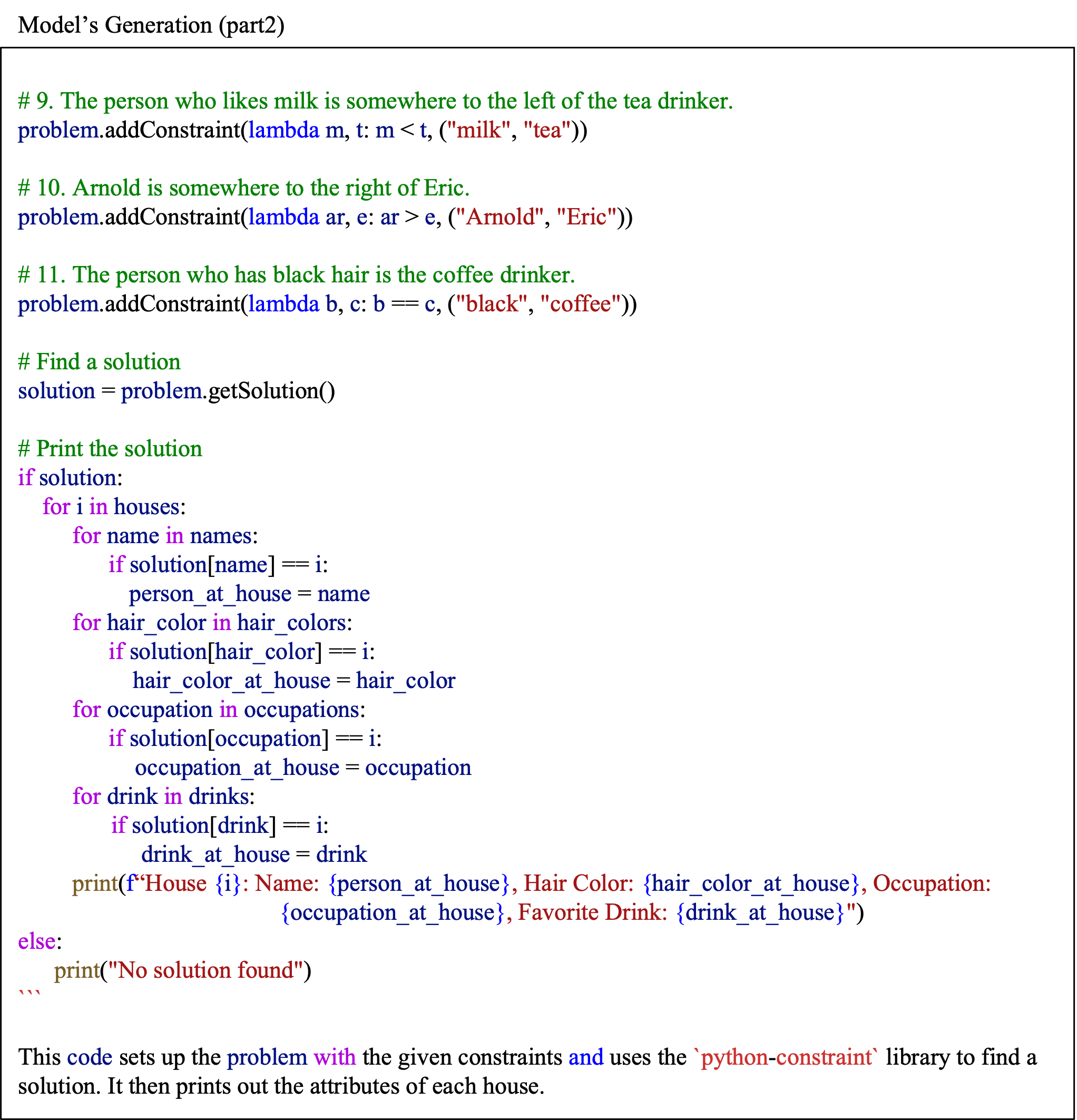}
  \caption{Generation from GPT-4o with Python on ZebraLogic (Part1)} \label{eneration-zebralogic-python-part2}
  \end{figure*}

   \begin{figure*}[!ht]
  \centering
    \includegraphics[scale=0.9]{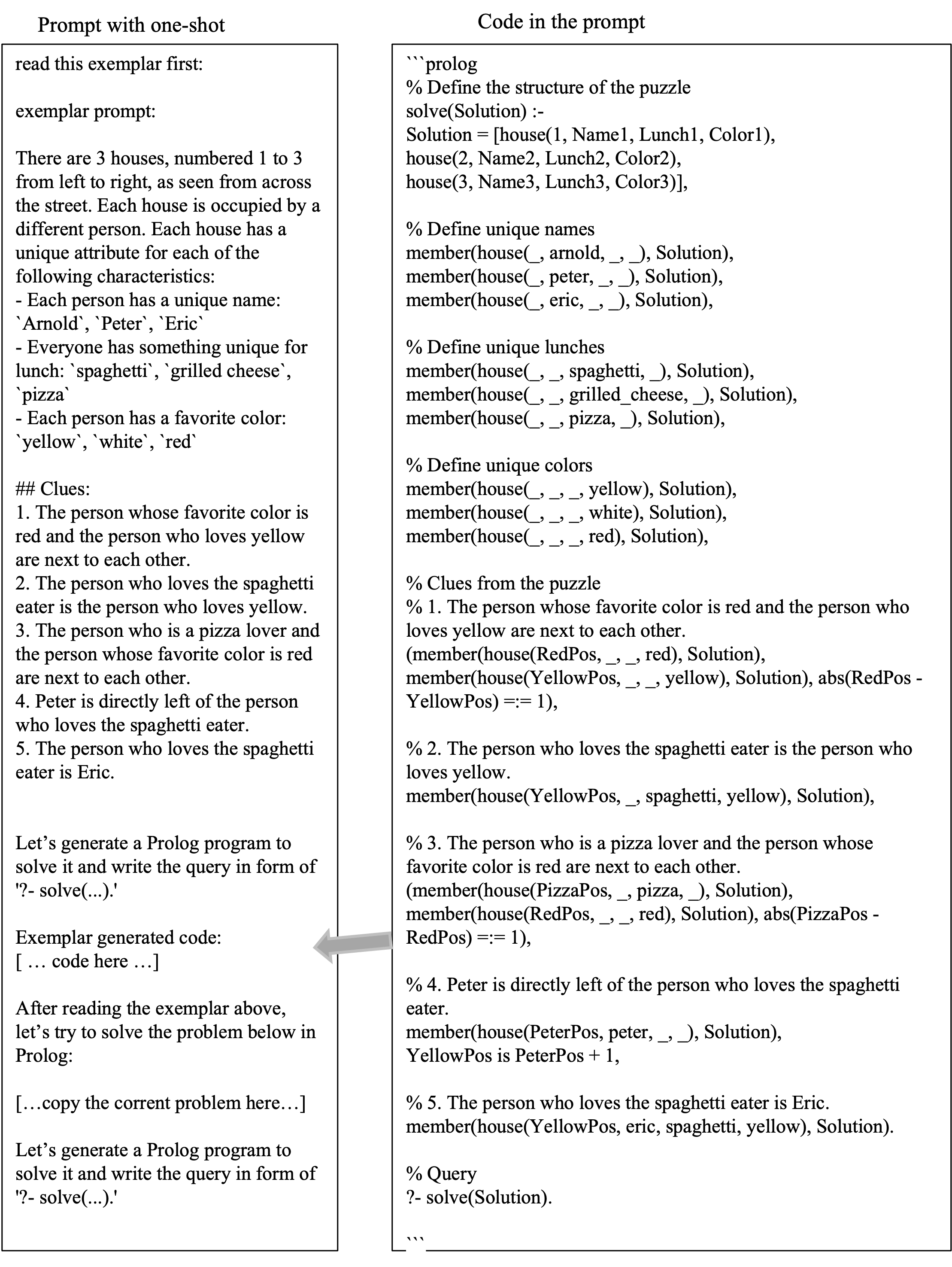}
  \caption{Prompt with a declarative Prolog exemplar for ZebraLogic} 
  \label{prompt-zebralogic-prolog}
  \end{figure*}

    \begin{figure*}[!ht]
  \centering
    \includegraphics[scale=0.83]{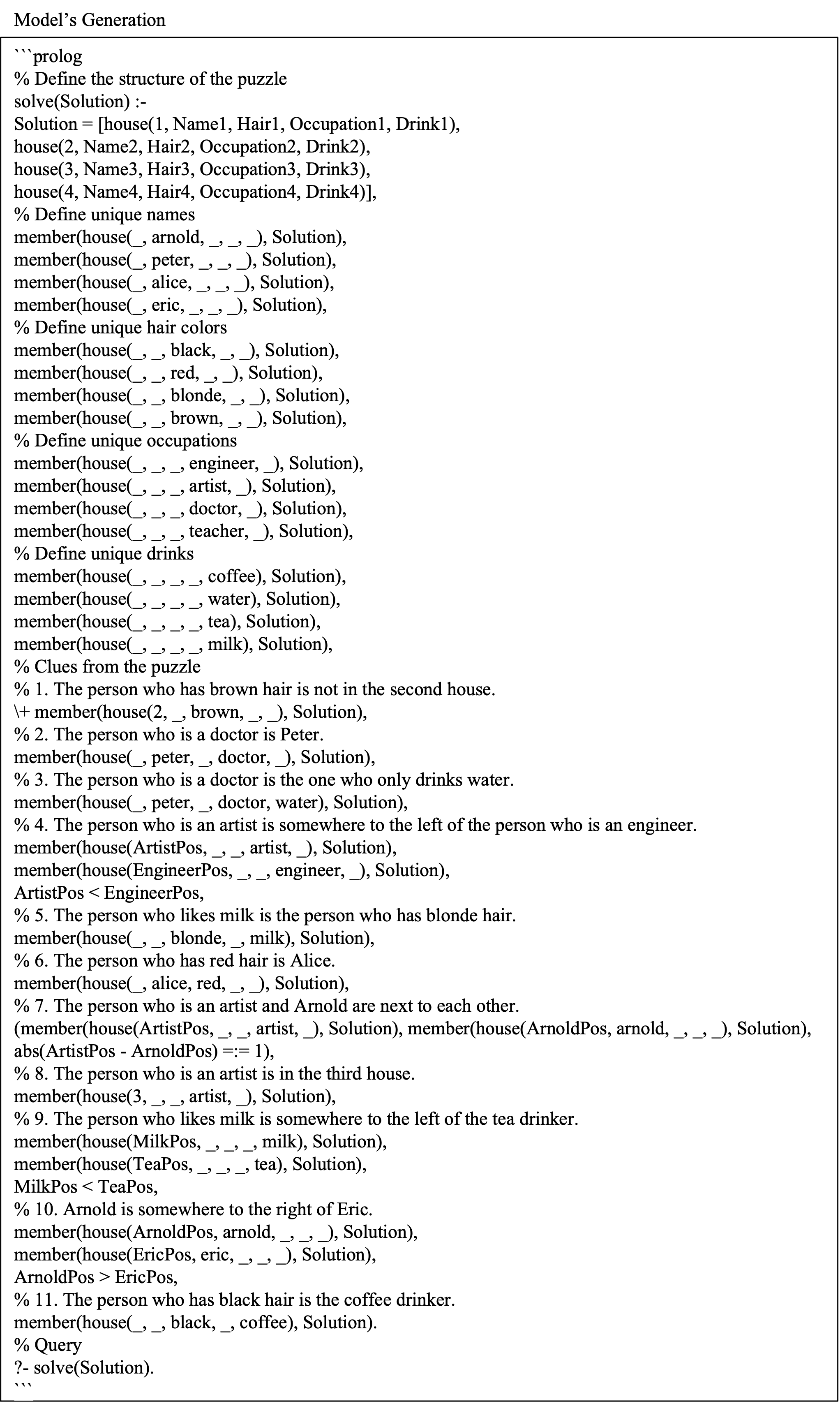}
  \caption{Generation from GPT-4o with Prolog on ZebraLogic} \label{generation-zebralogic-prolog}
  \end{figure*}

   \begin{figure*}[!ht]
  \centering
    \includegraphics[scale=1]{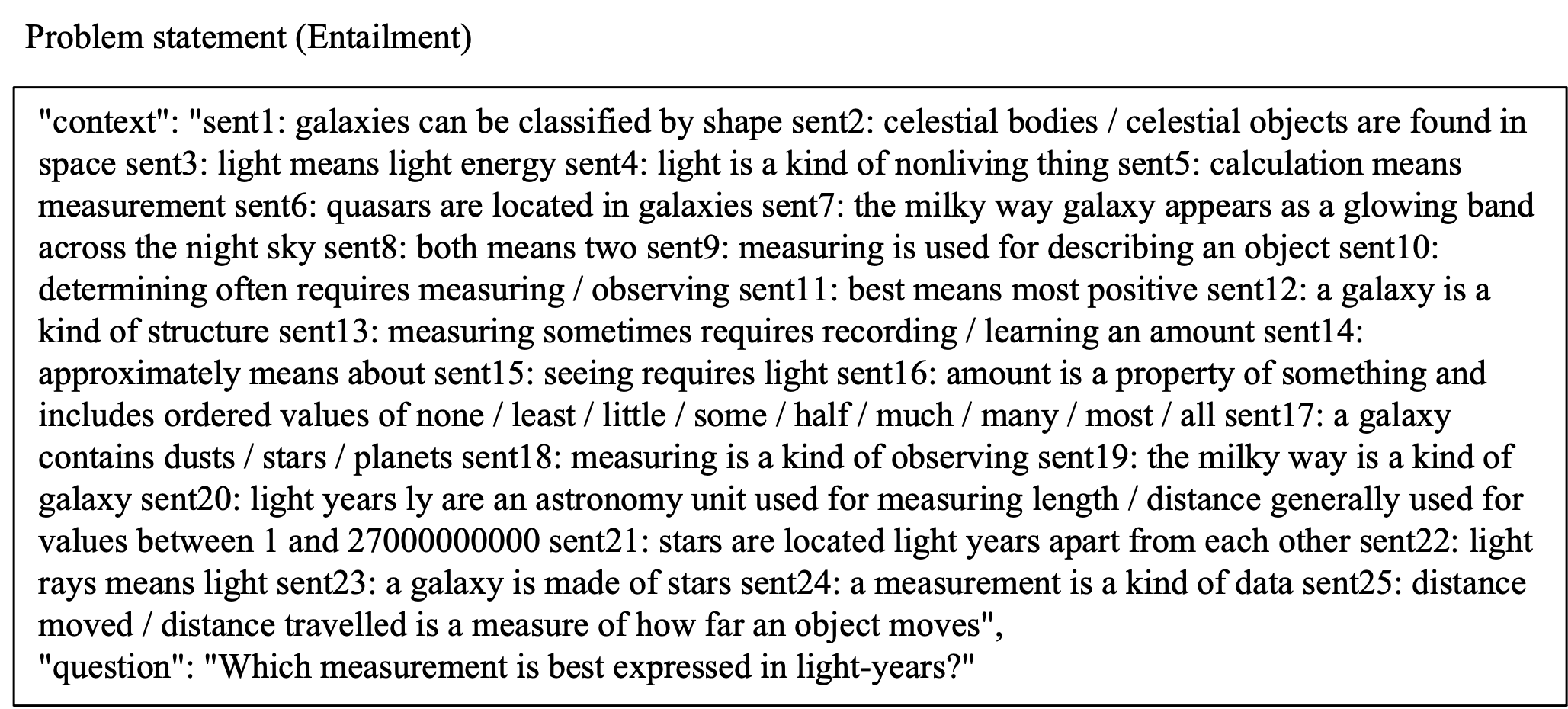}
  \caption{A problem in EntailmentBank} 
  \label{entailment-problem}
  \end{figure*}

\begin{figure*}[!ht]
  \centering
    \includegraphics[scale=1]{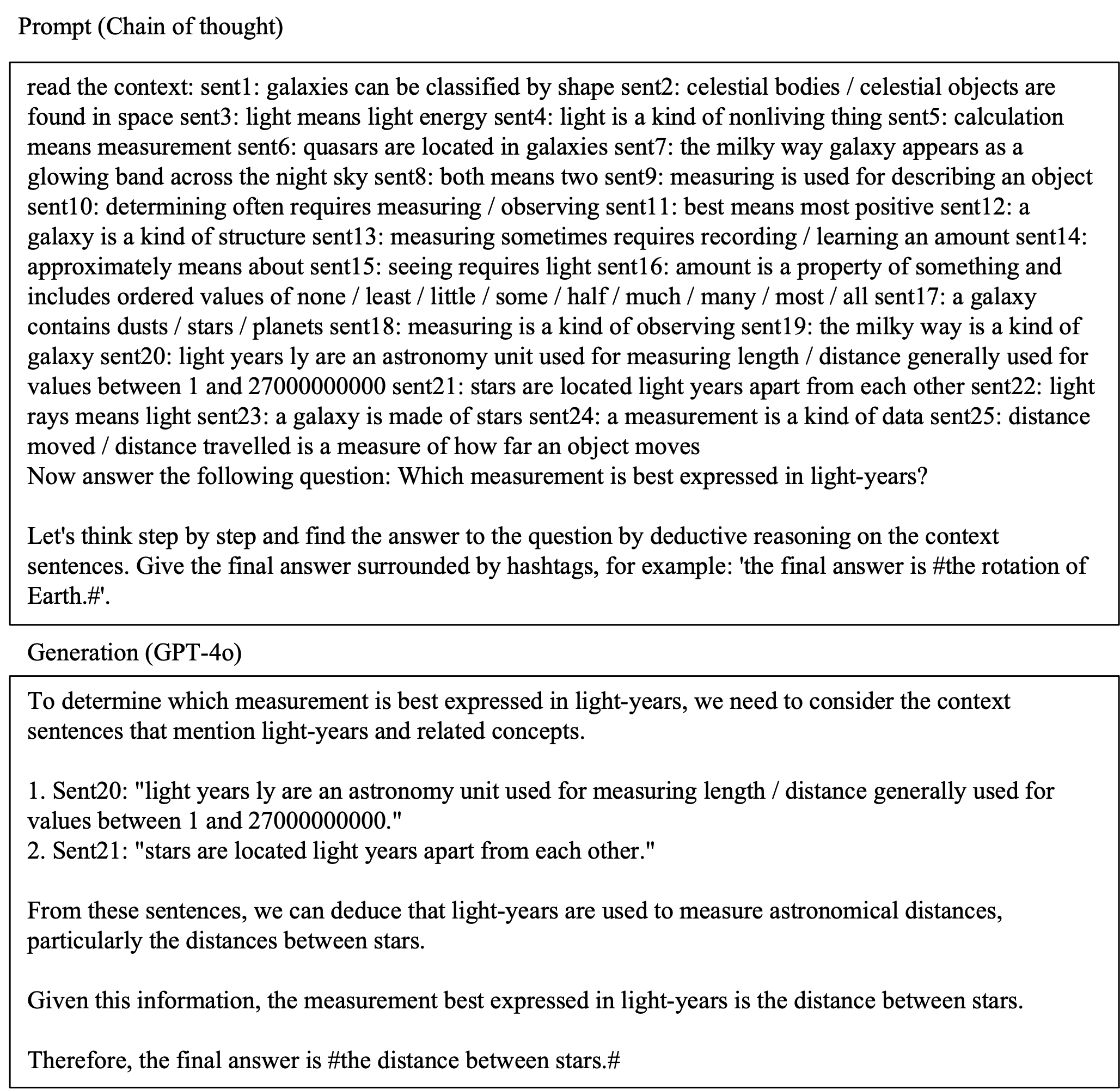}
  \caption{A Chain-Of-Thoughts Prompt for EntailmentBank} \label{entailment-prompt-generation-cot}
  \end{figure*}

\begin{figure*}[!ht]
  \centering
    \includegraphics[scale=1]{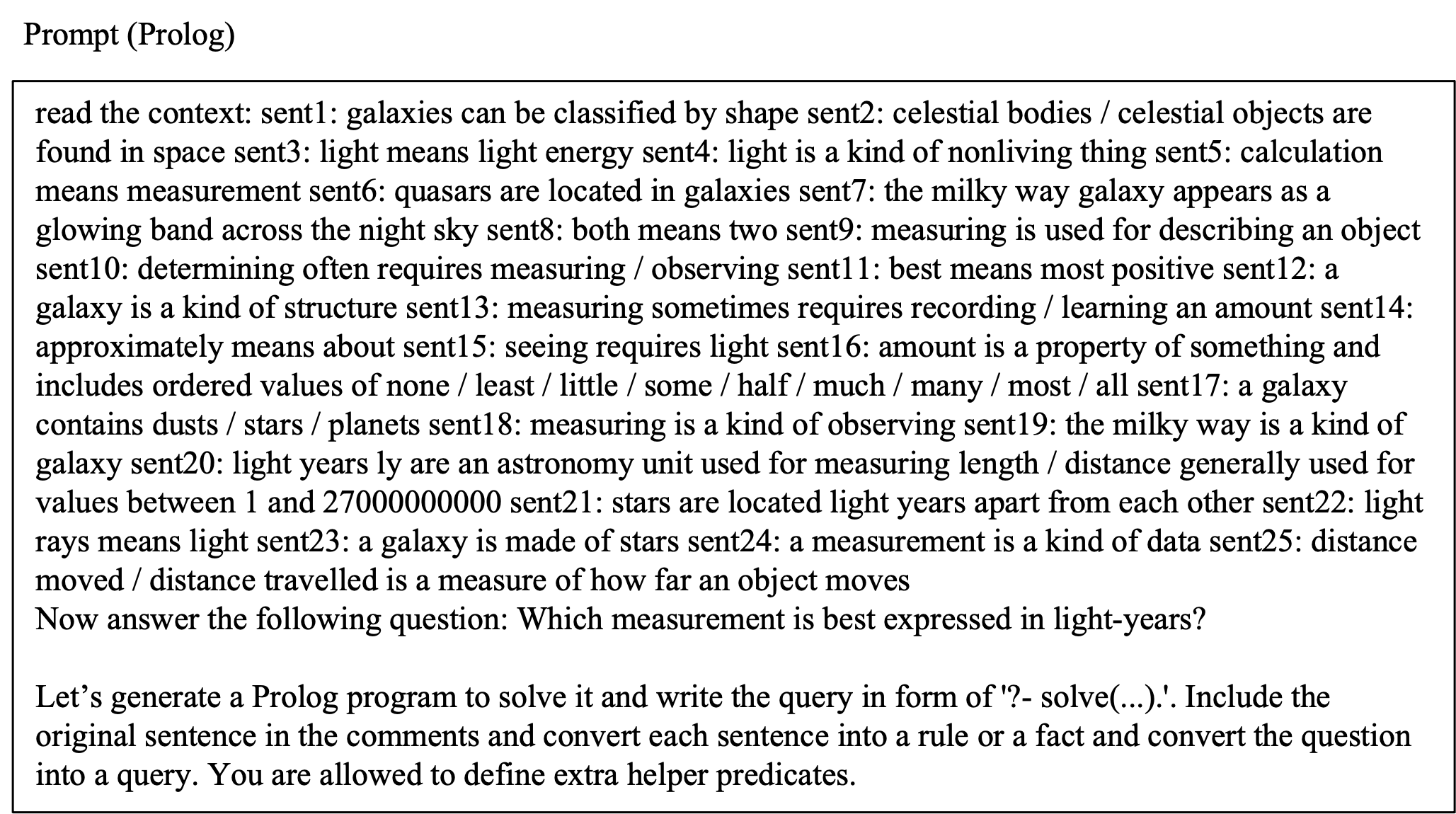}
  \caption{Prompt with a declarative Prolog exemplar for EntailmentBank} \label{entailment-prompt-prolog}
  \end{figure*}
  
\begin{figure*}[!ht]
  \centering
    \includegraphics[scale=0.85]{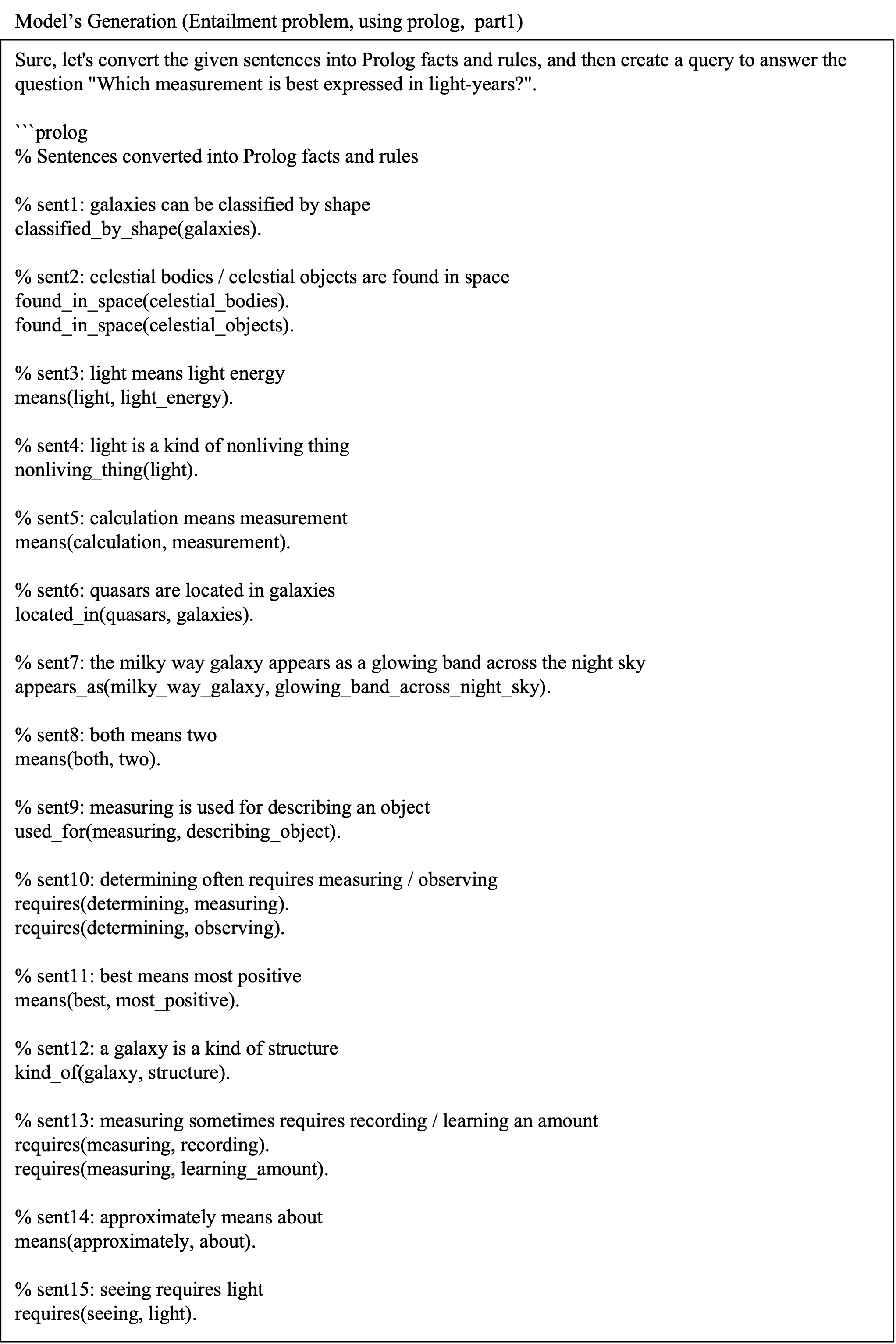}
  \caption{Generation from GPT-4o with Prolog on EntailmentBank (Part1)} \label{entailment-generation-prolog-part1}
  \end{figure*}

  \begin{figure*}[!ht]
  \centering
    \includegraphics[scale=0.85]{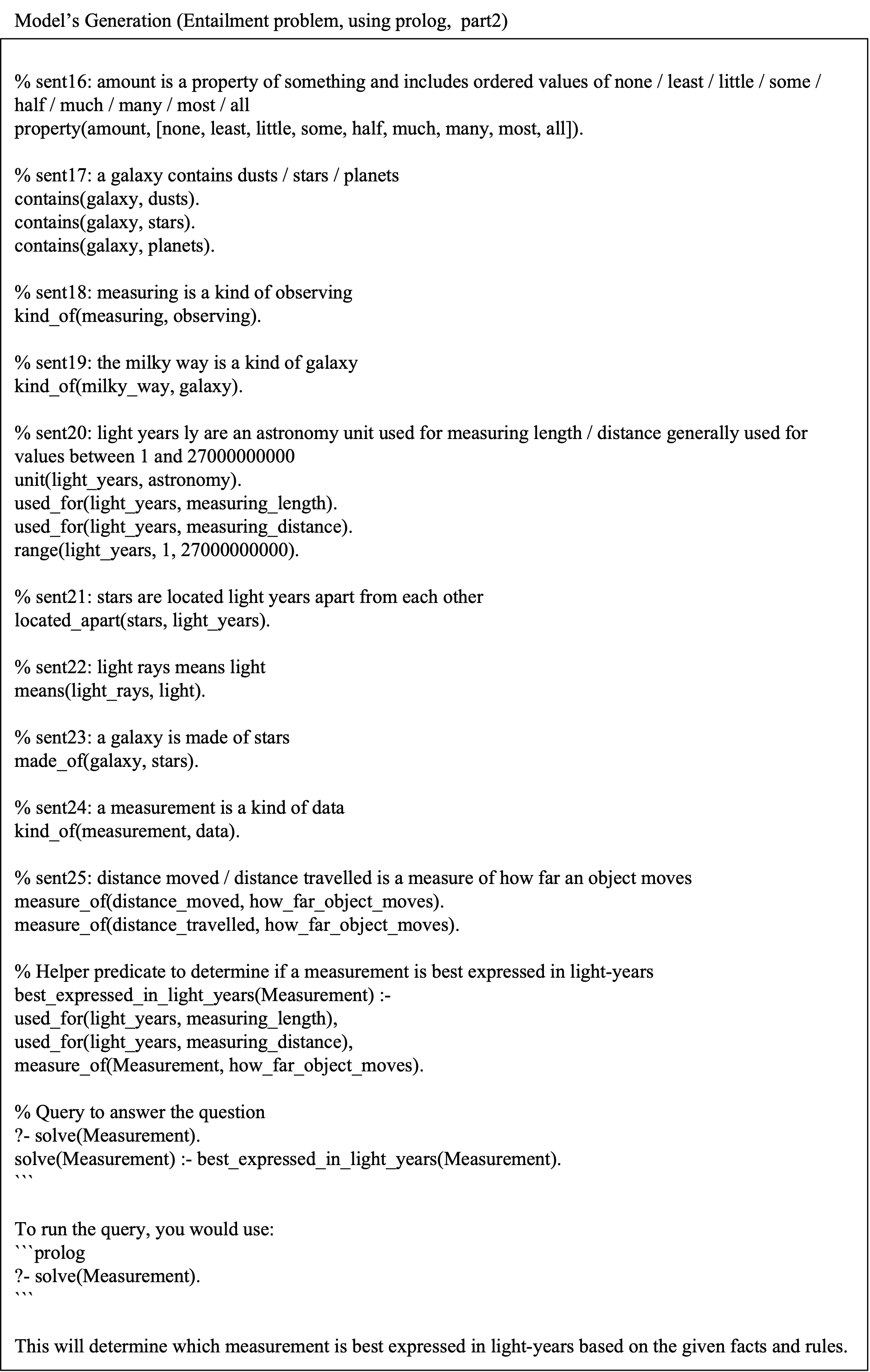}
  \caption{Generation from GPT-4o with Prolog on EntailmentBank (Part2)} \label{entailment-generation-prolog-part2}
  \end{figure*}

\clearpage
\section{ZebraLogic Results}
\label{Appendix-ZebraLogic-Results}


\begin{table}[!ht]
\centering
\subfloat[Using GPT-4o with Python code (one-shot)]{
    \begin{tabular}{c|ccccc}
    \toprule
    \multirow{2}{*}{\textbf{Houses}} & \multicolumn{5}{c}{\textbf{Number of Features}} \\
     & \textbf{2} & \textbf{3} & \textbf{4} & \textbf{5} & \textbf{6} \\
    \midrule
    \textbf{2} & 100 & 100 & 100 & 97.5 & 97.5 \\
    \textbf{3} & 92.5 & 87.5  & 90 & 87.5 & 72.5 \\
    \textbf{4} & 100 & 75 & 75 & 75 & 65 \\
    \textbf{5} & 90 & 70 & 62.5 & 42.5 & 30 \\
    \textbf{6} & 87.5 & 50 & 17.5 & 15.0 & 12.5 \\
    \bottomrule
    \end{tabular}
    \label{tab:zebralogic-GPT-4o-python}
}
\subfloat[Using GPT-4o with Prolog code (one-shot). The bold zeros indicate that the executor failed to return results due to a stack space overflow.]{
    \begin{tabular}{c|ccccc}
    \toprule
    \multirow{2}{*}{\textbf{Houses}} & \multicolumn{5}{c}{\textbf{Number of Features}} \\
     & \textbf{2} & \textbf{3} & \textbf{4} & \textbf{5} & \textbf{6} \\
    \midrule
    \textbf{2} & 100 & 100 & 100 & 97.5 & 97.5 \\
    \textbf{3} & 92.5 & 90.0  & 90.0 & 77.5 & 65.0 \\
    \textbf{4} & 97.5 & 80.0 & 72.5 & 70.0 & \textbf{0} \\
    \textbf{5} & 85.0 & 75 & \textbf{0} & \textbf{0} & \textbf{0} \\
    \textbf{6} & 80.0 & \textbf{0} & \textbf{0} & \textbf{0} & \textbf{0} \\
    \bottomrule
    \end{tabular}
    \label{tab:zebralogic-GPT-4o-prolog}
}\\
\subfloat[CodeLlama13B with Python code (one-shot)]{
    \begin{tabular}{c|ccccc}
    \toprule
    \multirow{2}{*}{\textbf{Houses}} & \multicolumn{5}{c}{\textbf{Number of Features}} \\
     & \textbf{2} & \textbf{3} & \textbf{4} & \textbf{5} & \textbf{6} \\
    \midrule
    \textbf{2} & 97.5 & 85.0 & 75.0 & 67.5 & 77.5 \\
    \textbf{3} & 57.5 & 60.0  & 42.5 & 40.0 & 25.0 \\
    \textbf{4} & 40 & 40 & 22.5 & 25 & 15 \\
    \textbf{5} & 40 & 22.5 & 7.5 & 5 & 0.15 \\
    \textbf{6} & 7.5 & 12.5 & 0 & 0 & 0 \\
    \bottomrule
    \end{tabular}
    \label{tab:zebralogic-cl13b-python}
}
\subfloat[CodeLlama13B with Prolog code (one-shot). The bold zeros indicate that the executor failed to return results due to a stack space overflow.]{
    \begin{tabular}{c|ccccc}
    \toprule
    \multirow{2}{*}{\textbf{Houses}} & \multicolumn{5}{c}{\textbf{Number of Features}} \\
     & \textbf{2} & \textbf{3} & \textbf{4} & \textbf{5} & \textbf{6} \\
    \midrule
    \textbf{2} & 85 & 75 & 67.5 & 62.5 & 42.5 \\
    \textbf{3} & 50 & 55  & 32.5 & 25 & 12.5 \\
    \textbf{4} & 42.5 & 32.5 & 22.5 & 17.5 & \textbf{0} \\
    \textbf{5} & 35 & 12.5 & \textbf{0} & \textbf{0} & \textbf{0} \\
    \textbf{6} & 15 & \textbf{0} & \textbf{0} & \textbf{0} & \textbf{0} \\
    \bottomrule
    \end{tabular}
    \label{tab:zebralogic-cl13b-prolog}
}\\
\caption{Accuracies of symbolic-solver-integrated methods on ZebraLogic puzzles}
\label{tab:zebralogic-cl13b}
\end{table}


\begin{table}[h!]
\centering
\subfloat[Using GPT-4o]{
    \begin{tabular}{c|ccccc}
    \toprule
    \multirow{2}{*}{\textbf{Houses}} & \multicolumn{5}{c}{\textbf{Number of Features}} \\
     & \textbf{2} & \textbf{3} & \textbf{4} & \textbf{5} & \textbf{6} \\
    \midrule
    \textbf{2} & 100 & 100 & 90 & 82.5 & 70 \\
    \textbf{3} & 65 & 37.5  & 25 & 30 & 12.5 \\
    \textbf{4} & 45 & 20 & 12.5 & 0 & 2.5 \\
    \textbf{5} & 12.5 & 0 & 2.5 & 0 & 0 \\
    \textbf{6} & 7.5 & 0 & 0 & 0 & 0 \\
    \bottomrule
    \end{tabular}
    \label{tab:zebralogic-GPT-4o}
}
\subfloat[Using Claude-3.5-Sonnet]{
    \begin{tabular}{c|ccccc}
    \toprule
    \multirow{2}{*}{\textbf{Houses}} & \multicolumn{5}{c}{\textbf{Number of Features}} \\
     & \textbf{2} & \textbf{3} & \textbf{4} & \textbf{5} & \textbf{6} \\
    \midrule
    \textbf{2} & 100 & 100 & 90 & 95 & 90 \\
    \textbf{3} & 75 & 62.5  & 37.5 & 30 & 17.5 \\
    \textbf{4} & 55 & 37.5 & 10 & 2.5 & 2.5 \\
    \textbf{5} & 15 & 10 & 0 & 0 & 0 \\
    \textbf{6} & 5 & 0 & 0 & 0 & 0 \\
    \bottomrule
    \end{tabular}
    \label{tab:zebralogic-claude}
}\\
\caption{Accuracies of CoT on ZebraLogic puzzles}
\label{tab:zebralogic-cot-baselines}
\end{table}


\clearpage
\section{Failure modes}
\label{appendix-failure-modes}

 \begin{figure}[!ht]
  \centering
    \includegraphics[scale=0.8]{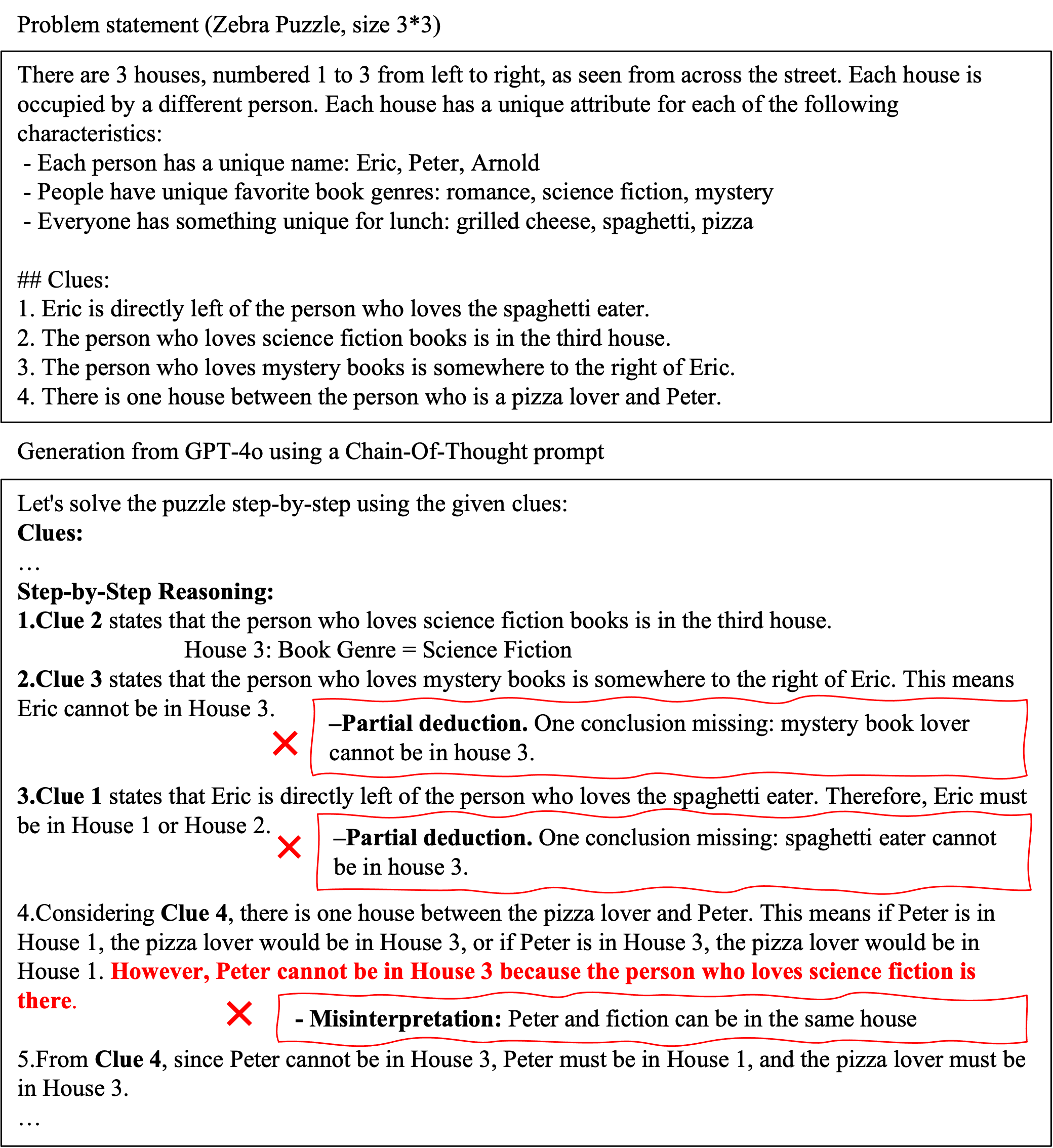}
  \caption{GPT-4o makes errors} \label{ZebraLogic-partial-deduction-misinterpretation}
  \end{figure}

   \begin{figure*}[!ht]
  \centering
    \includegraphics[scale=0.8]{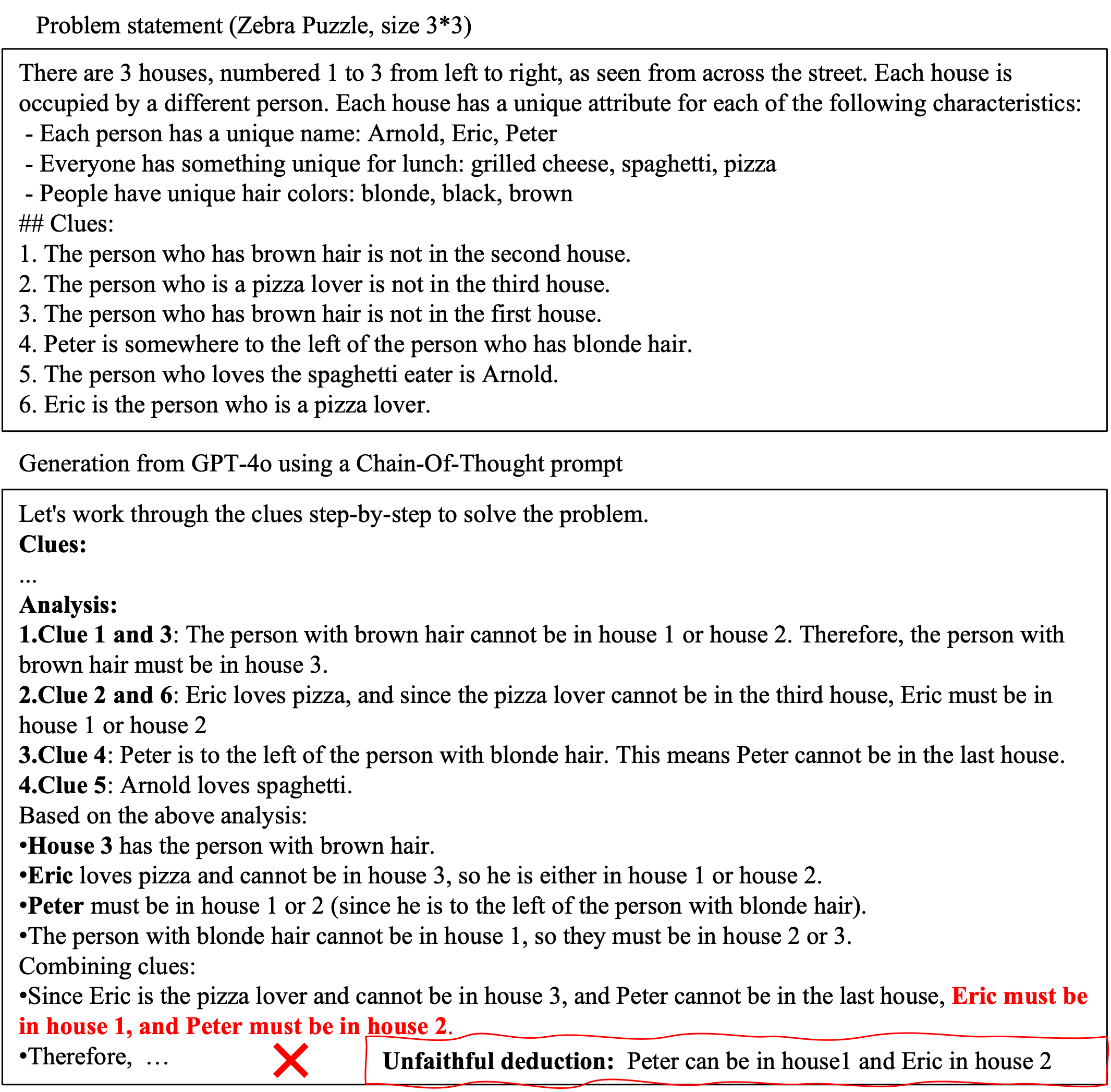}
  \caption{GPT-4o makes errors in ZebraLogic puzzles: unfaithful deduction} \label{ZebraLogic-unfaithful-deduction}
  \end{figure*}

   \begin{figure*}[!ht]
  \centering
    \includegraphics[scale=0.8]{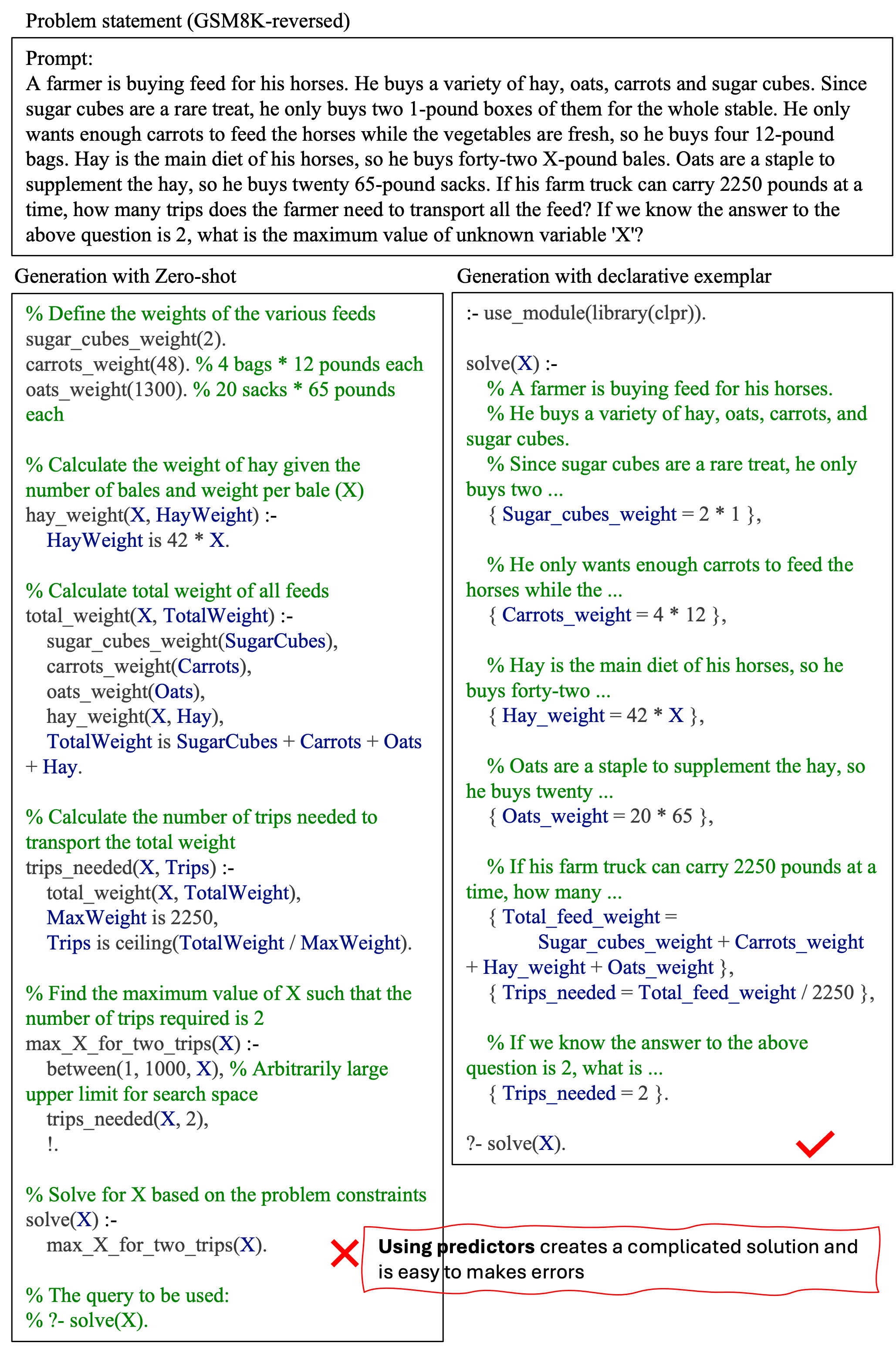}
  \caption{GPT-4o generates excessive predicates} \label{GSM-Reversed-using-predicators}
  \end{figure*}

 \begin{figure*}[!ht]
  \centering
    \includegraphics[scale=0.8]{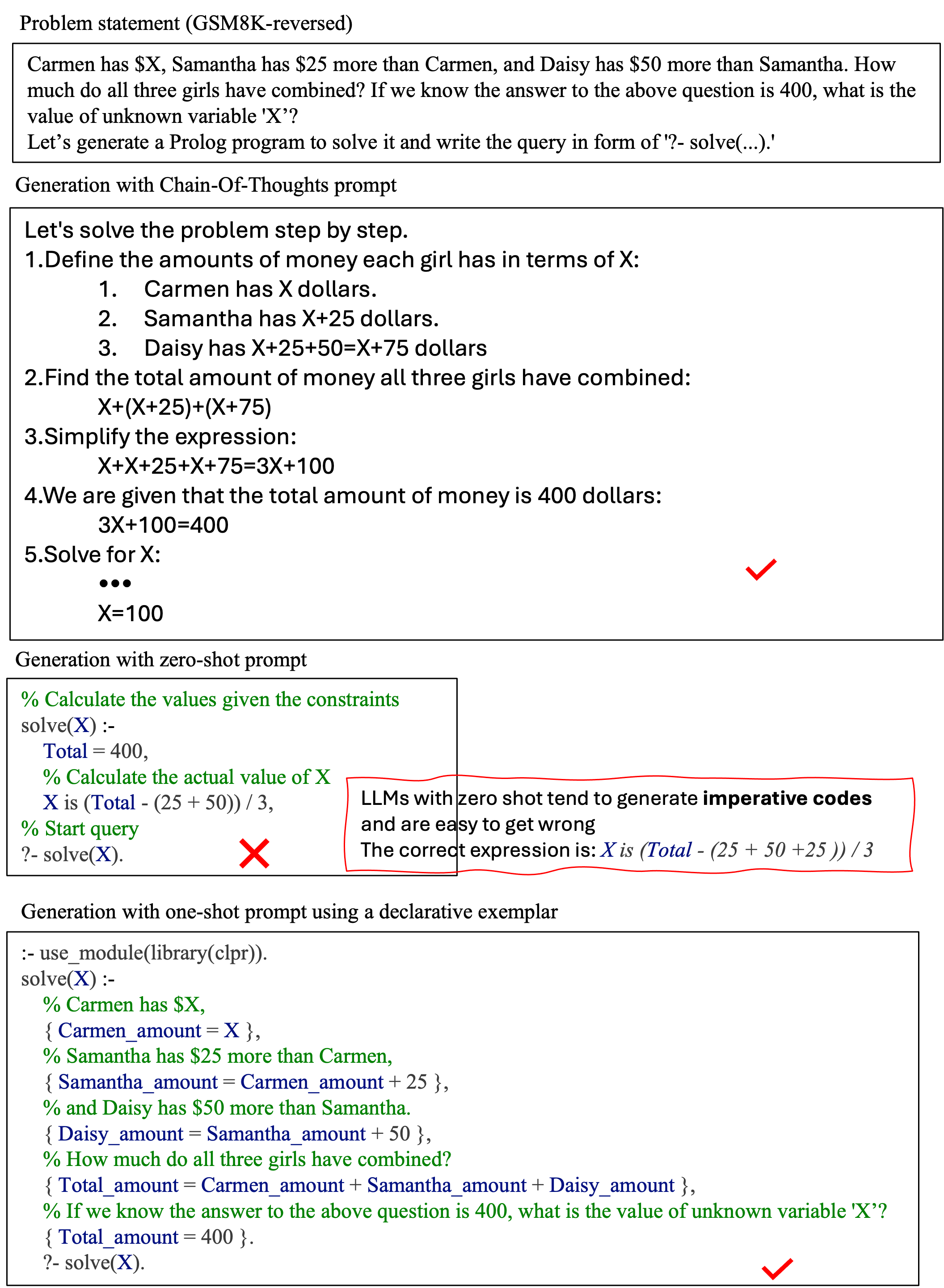}
  \caption{GPT-4o generates problematic imperative codes in GSM-Reversed} \label{failure-gsm-reversed-imperative-code}
  \end{figure*}

\clearpage
\section{Histogram Of Retries}
\label{Histogram Of Retries}
\begin{figure}[h!]
  \centering
\subfloat[Python code with zero-shot]{
    \includegraphics[scale=0.225]{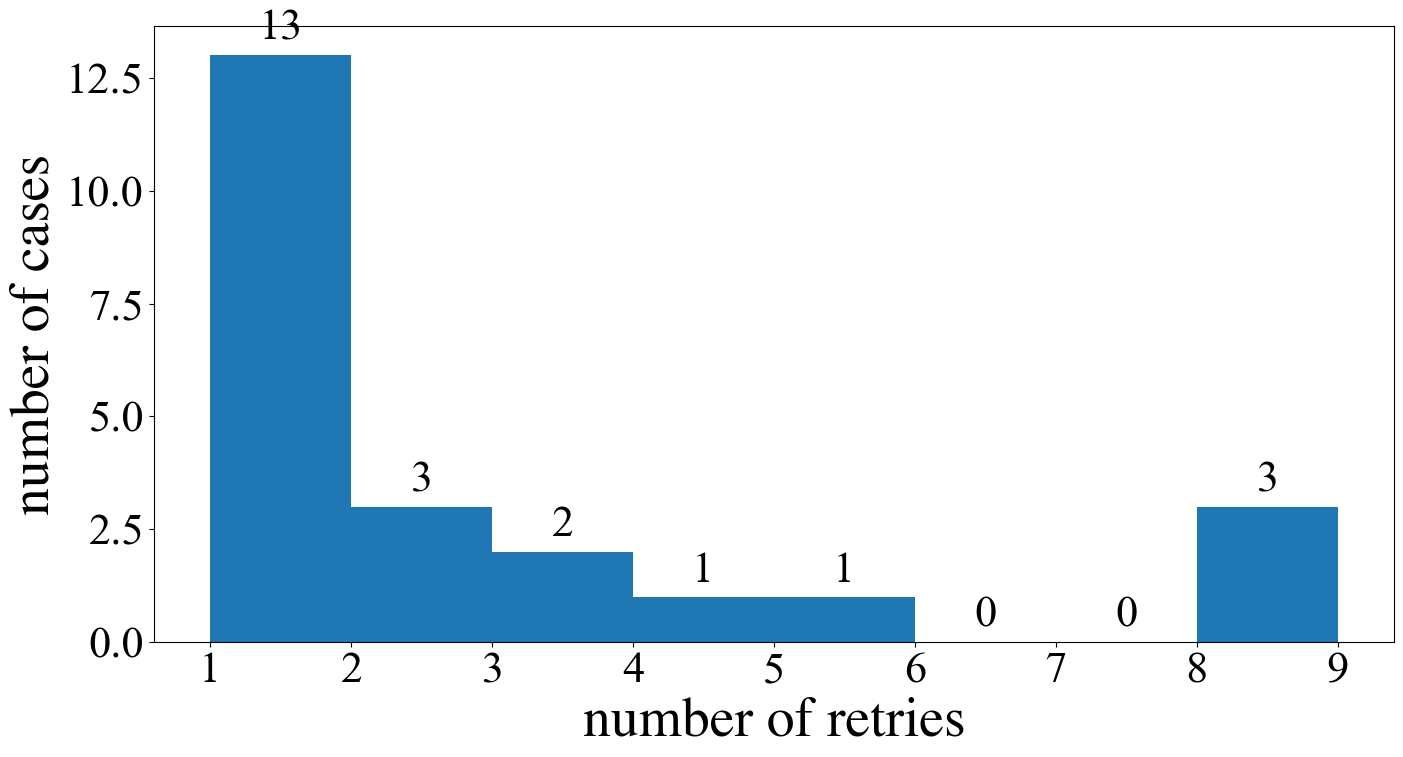}
    \label{gsm-hard-py-0s}
    }
  \subfloat[Python code with one-shot]{
    \includegraphics[scale=0.225]{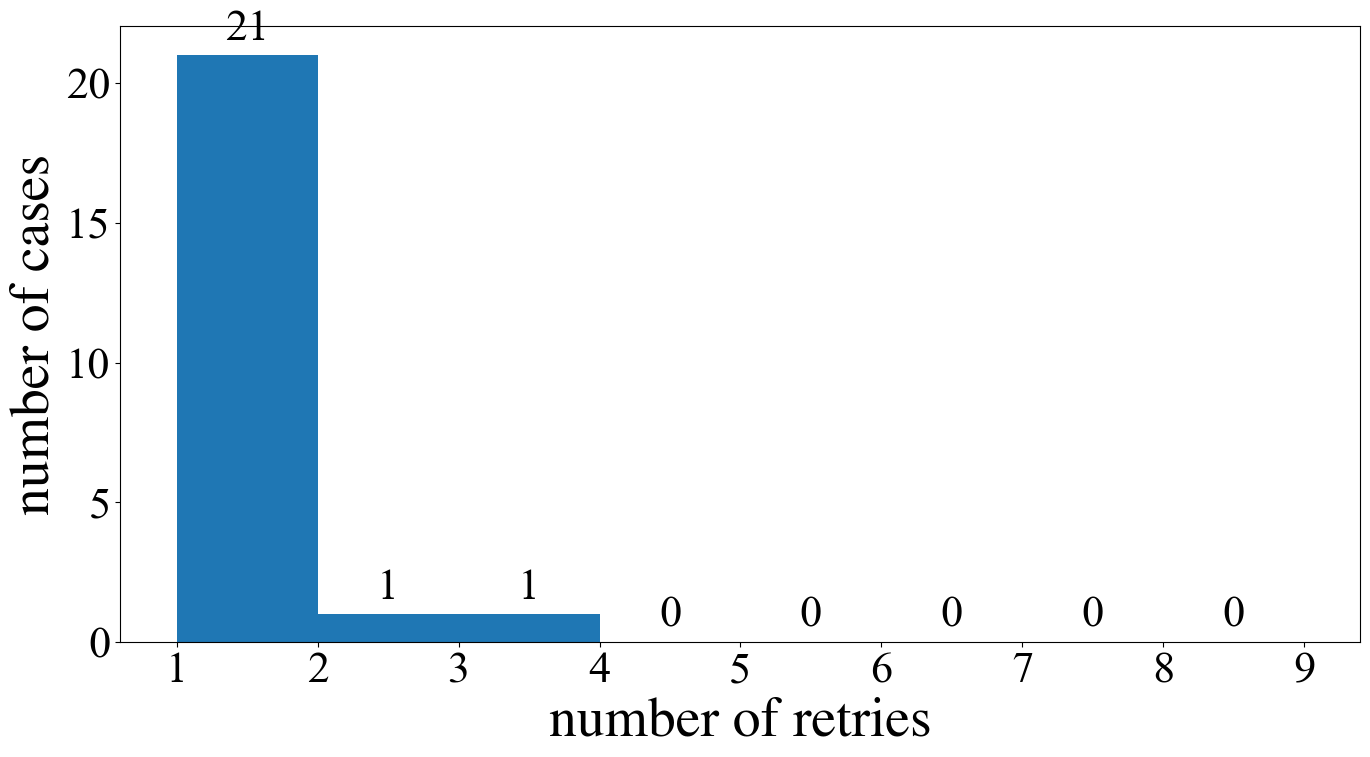}
    \label{gsm-hard-py-1s}
    }
  \caption{GSM-Hard using GPT-4o} \label{results-retries-gsm-hard-gpt4o}
\end{figure}

 \begin{figure}[h!]
  \centering
  
  \subfloat[Prolog code (Zero Shot)]{
    \includegraphics[scale=0.225]{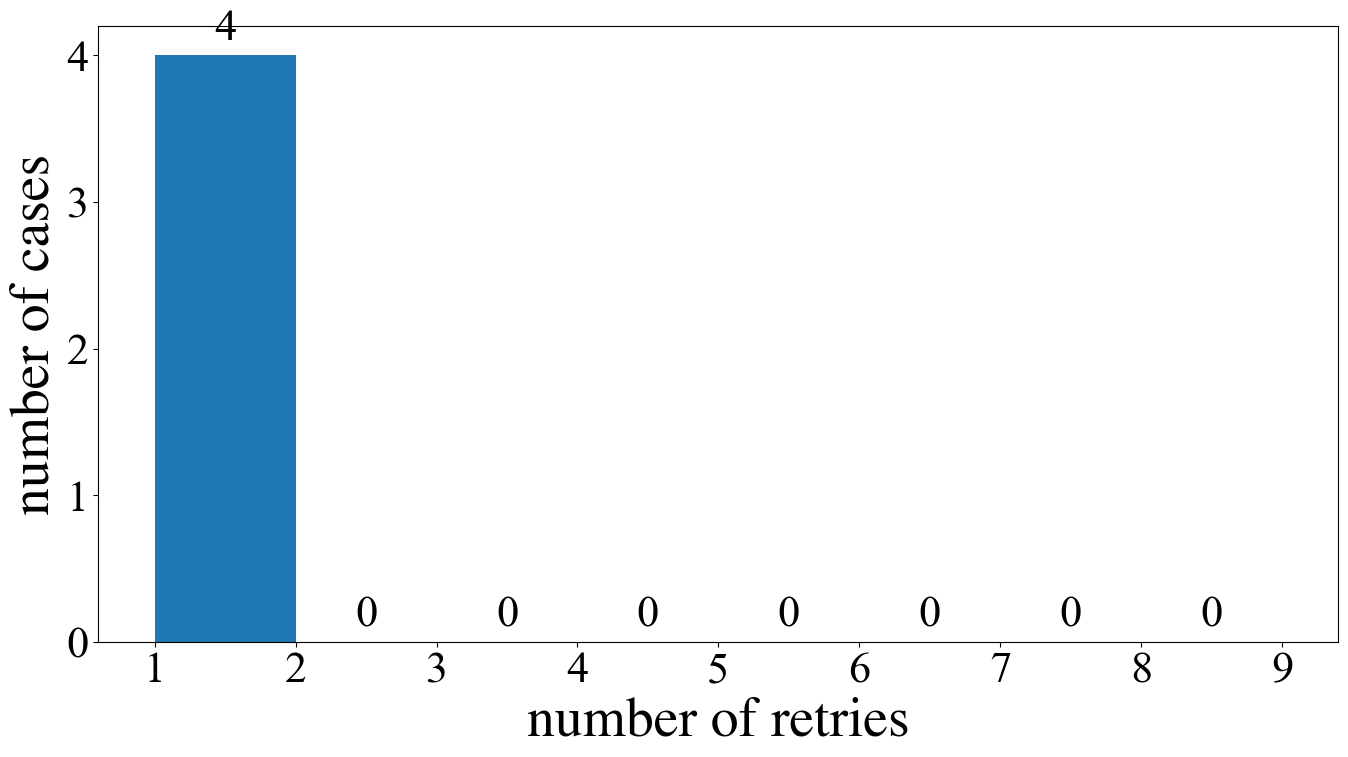}
    \label{ZebraLogic-GPT-4o-Prolog-Zero-Shot}
}
    \subfloat[Prolog code (One Shot)]{
    \includegraphics[scale=0.225]{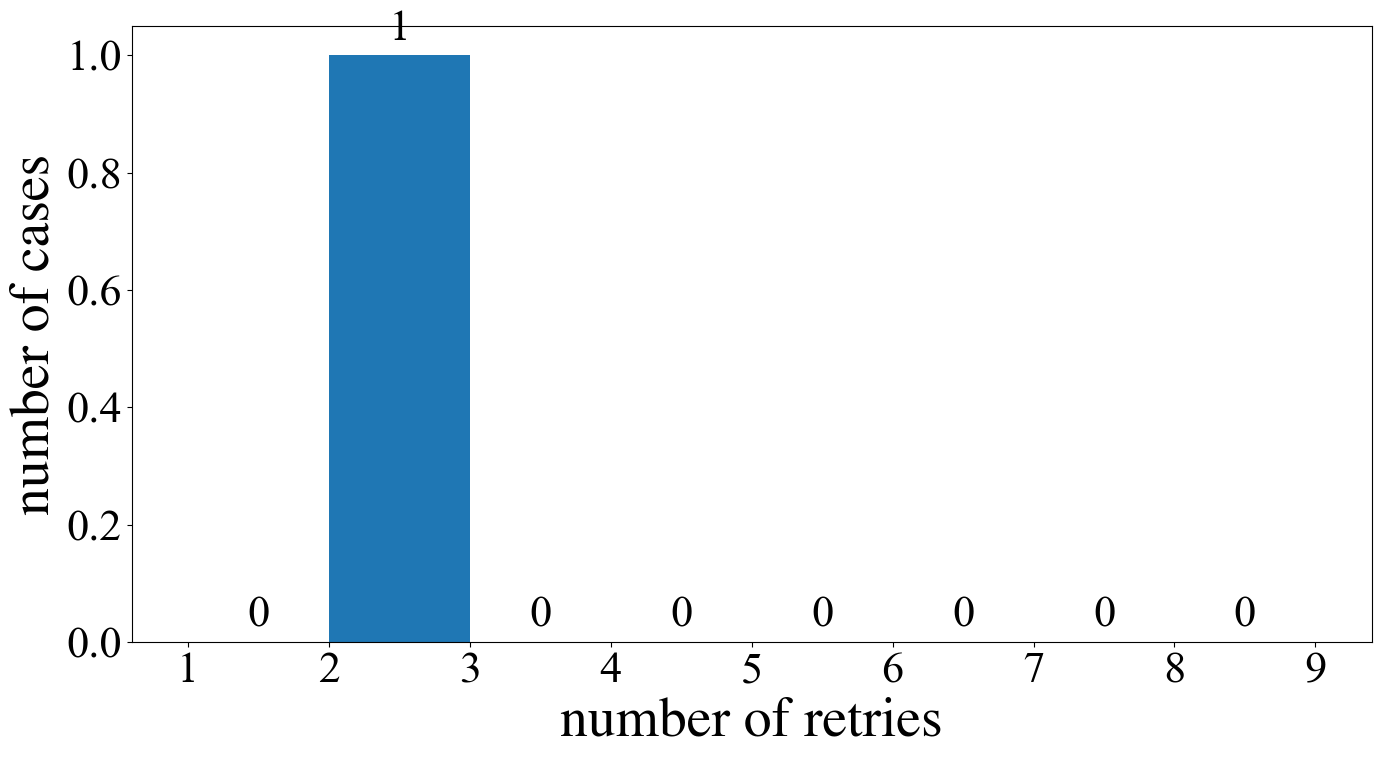}
    \label{ZebraLogic-GPT-4o-Prolog-One-Shot}
    } \\
  \subfloat[Python code (Zero Shot)]{
    \includegraphics[scale=0.225]{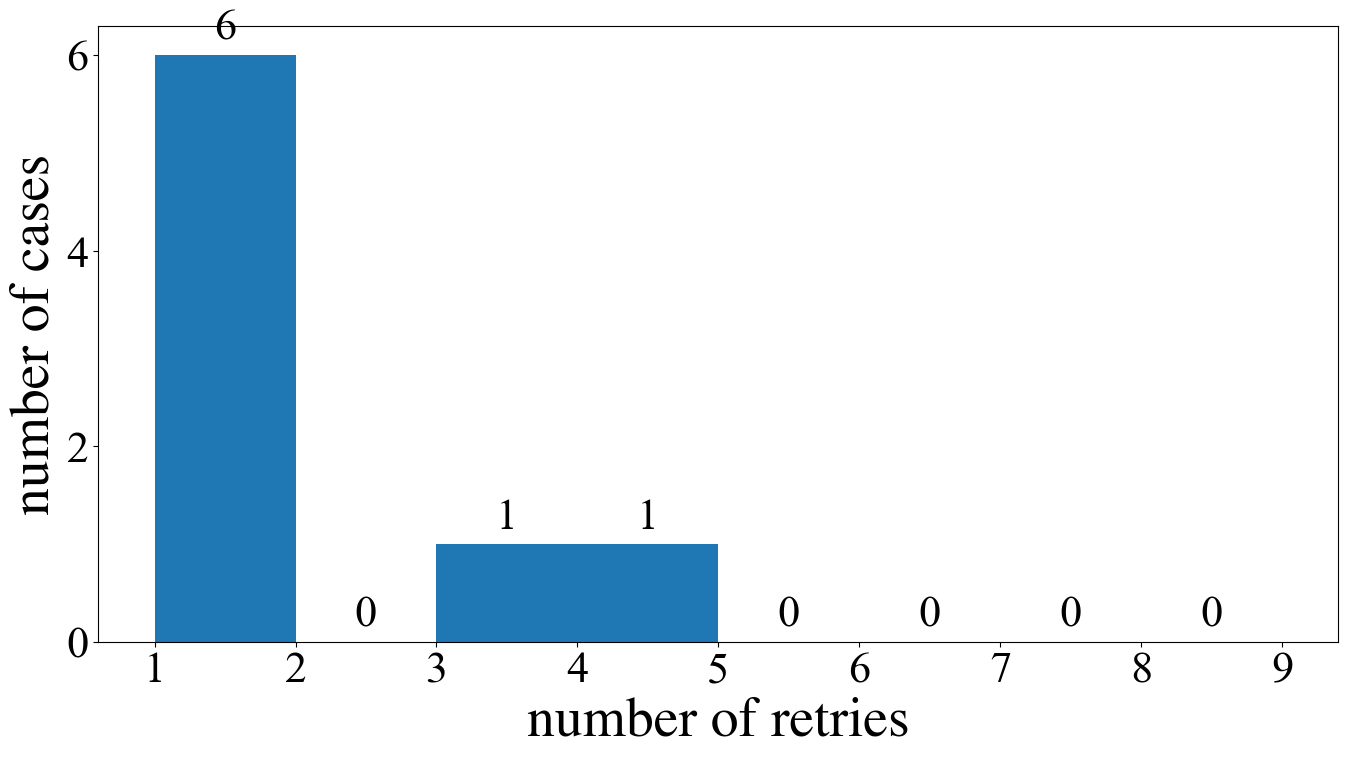}
    \label{ZebraLogic-GPT-4o-Python-Zero-Shot}
    }
  \subfloat[Python code (One Shot)]{
    \includegraphics[scale=0.225]{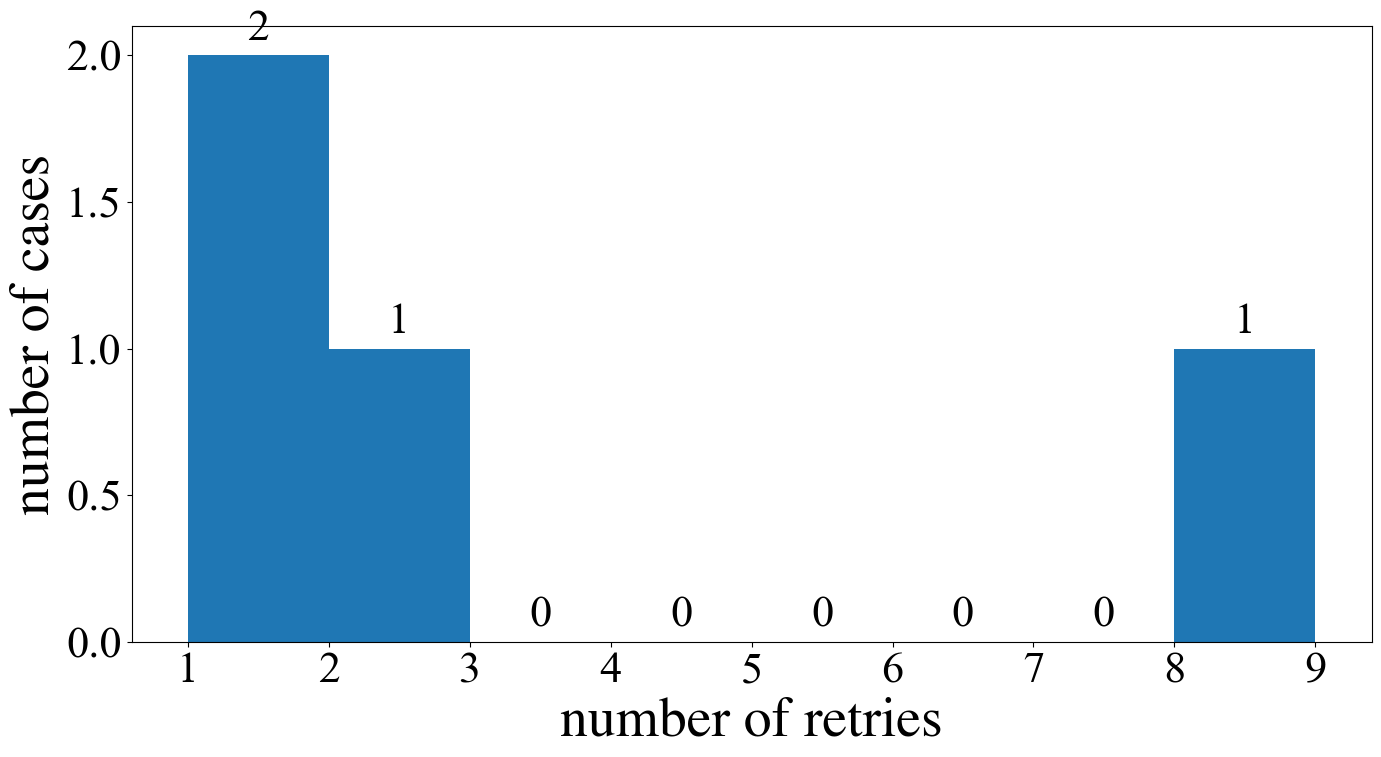}
    \label{ZebraLogic-GPT4o-Python-One-Shot}
    }
  \caption{GPT4o for ZebraLogic puzzles} \label{ZebraLogic-GPT-4o}
  \end{figure}

\begin{figure*}[!ht]
  \centering
  
  \subfloat[Prolog code (One Shot) for ZebraLogic]{
    \includegraphics[scale=0.225]{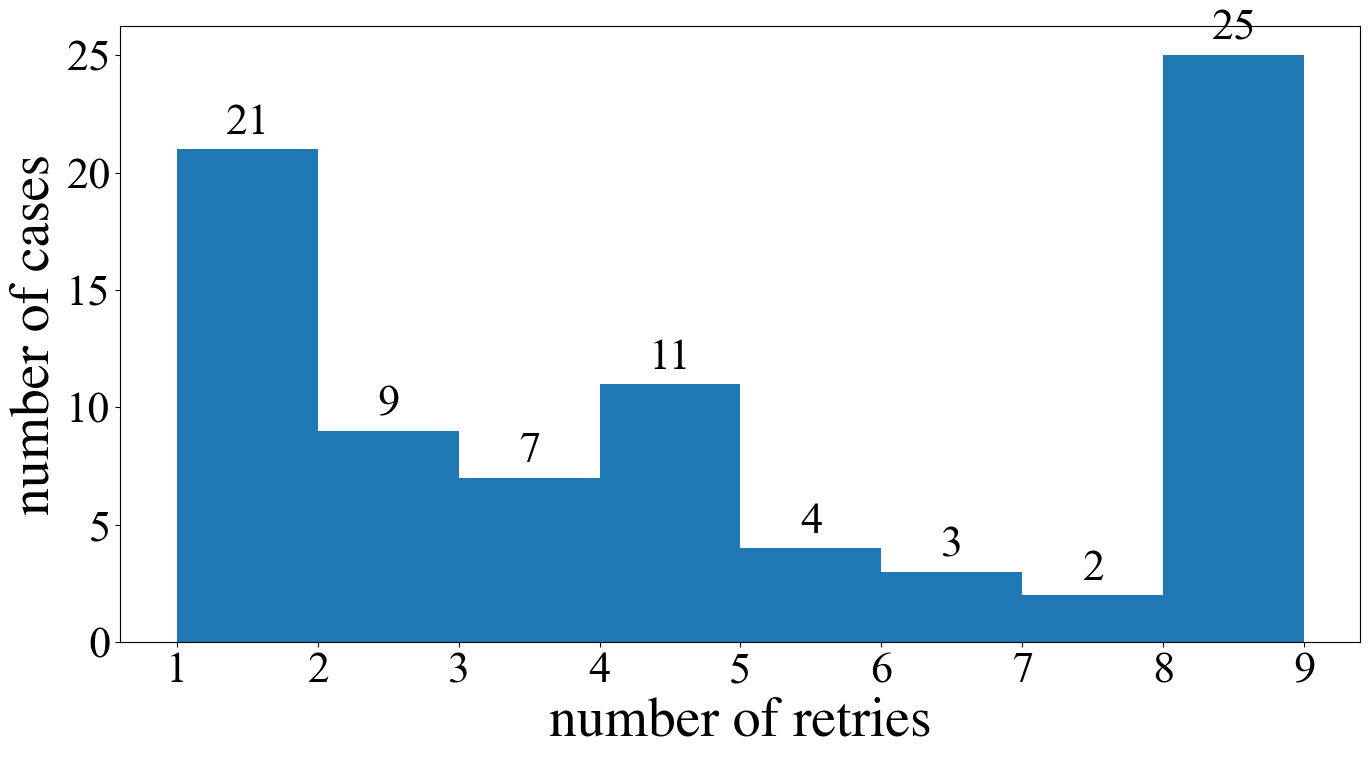}
    \label{ZebraLogic-CodeLlama13B-Prolog}
}
    \subfloat[Python code (One Shot) for ZebraLogic]{
    \includegraphics[scale=0.225]{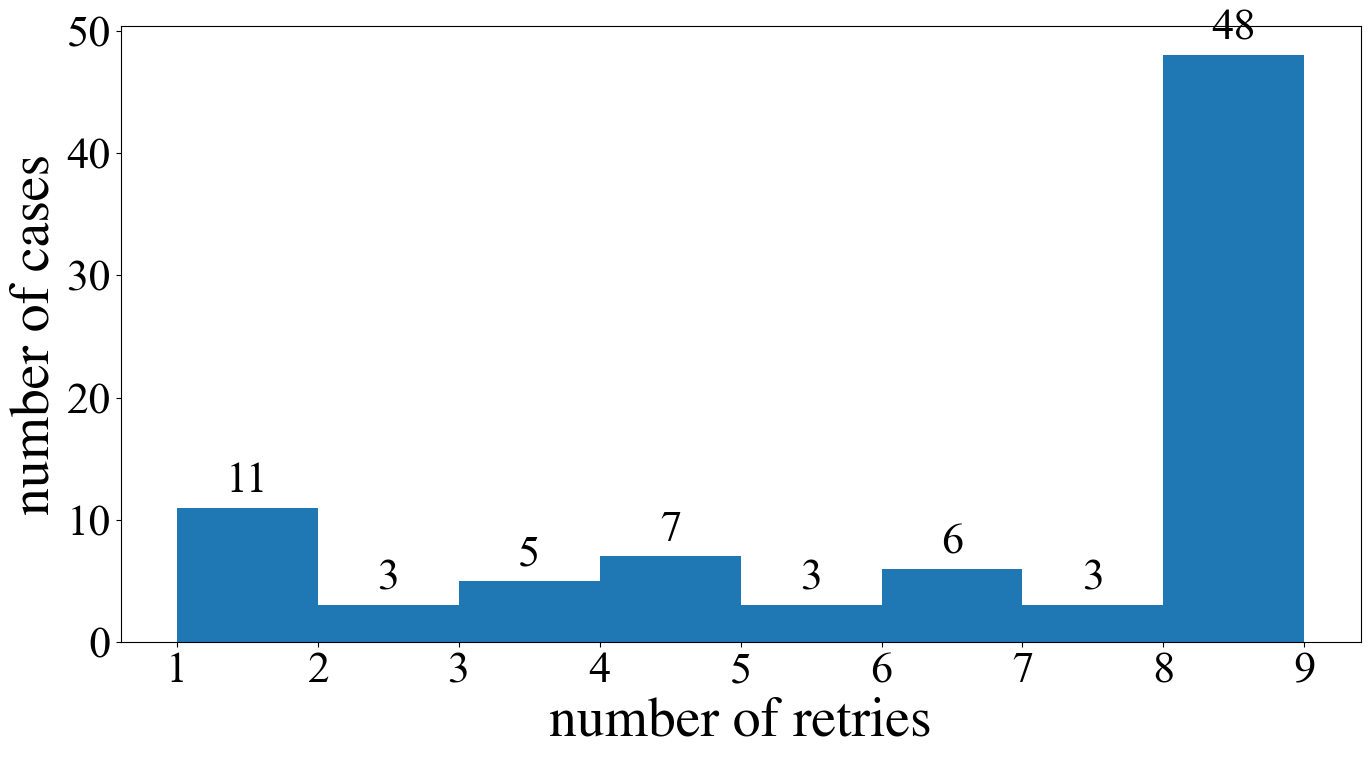}
    \label{ZebraLogic-CodeLlama13B-Python}
    }\\
    
    \subfloat[Prolog code (One Shot) for GSM-Hard]{
    \includegraphics[scale=0.225]{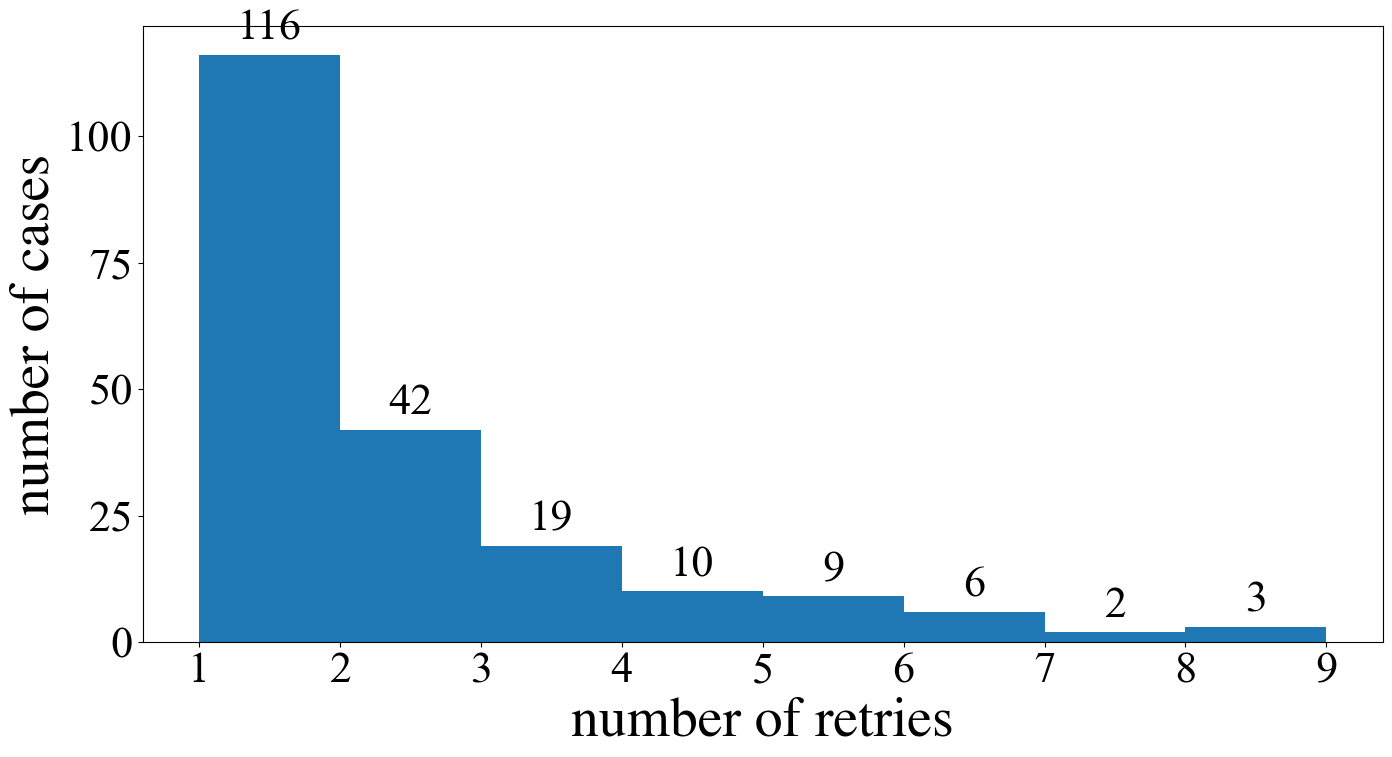}
    \label{GSM-Hard-CodeLlama13B-Prolog}
}
    \subfloat[Python code (One Shot) for GSM-Hard]{
    \includegraphics[scale=0.225]{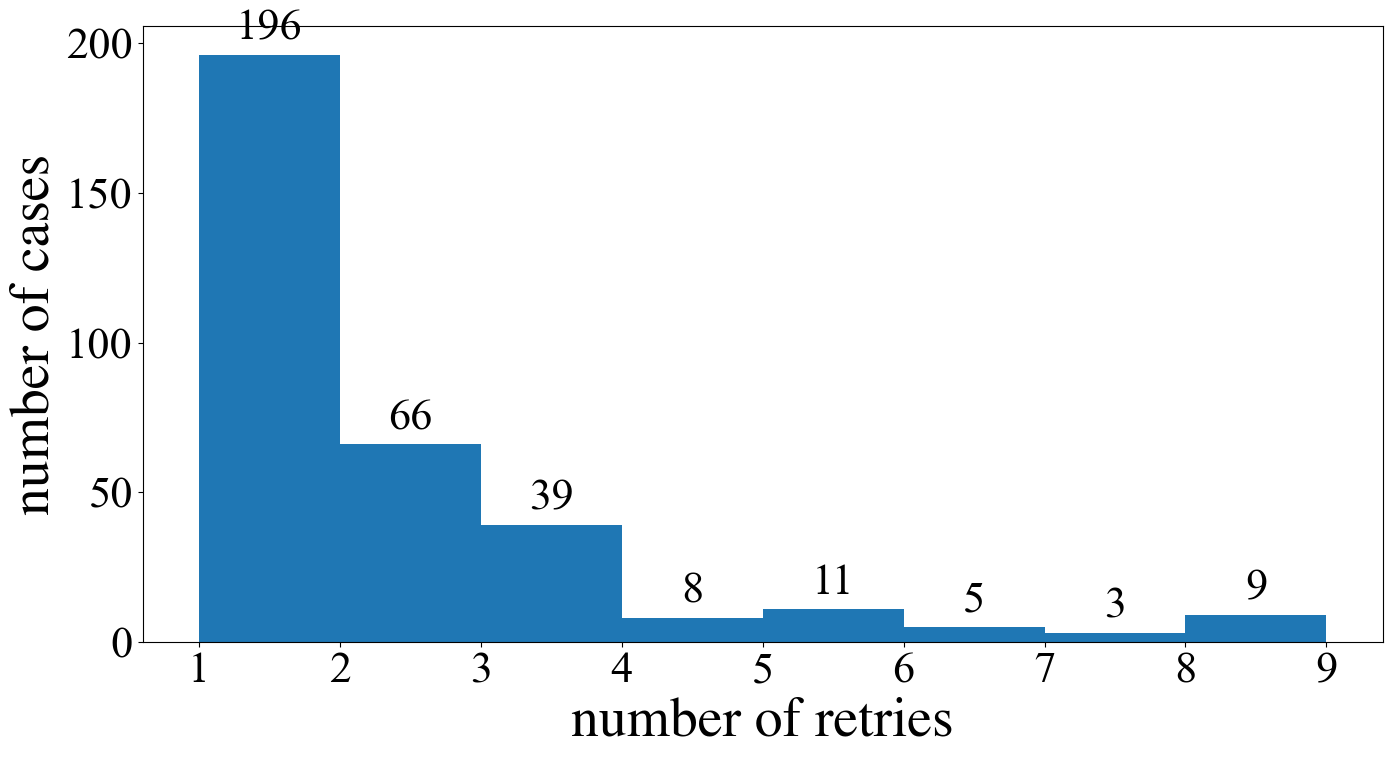}
    \label{GSM-Hard-CodeLlama13B-Python}
    }
  \caption{Investigate the retries: using CodeLlama13B for ZebraLogic puzzles and GSM-Hard} \label{ZebraLogic-CodeLlama13B}
  \end{figure*}

\clearpage


\end{document}